\documentclass{article}

\usepackage{arxiv}

\usepackage[utf8]{inputenc} 
\usepackage[T1]{fontenc}    
\usepackage{hyperref}       
\usepackage{url}            
\usepackage{booktabs}       
\usepackage{amsfonts}       
\usepackage{nicefrac}       
\usepackage{microtype}      
\usepackage{lipsum}		
\usepackage{graphicx}
\usepackage{natbib}
\usepackage{doi}
\usepackage{algorithm}
\usepackage{algorithmic}
\usepackage{amsmath}
\usepackage[english]{babel}

\title{Enhancing Interpretability of Sparse Latent Representations with Class Information}

\author{
	\textbf{Farshad Sangari Abiz}, \textbf{Reshad Hosseini}, \textbf{Babak N. Araabi} \\
	School of Electrical and Computer Engineering,\\
	University College of Engineering,\\
	University of Tehran, Tehran, Iran\\
	\texttt{\{f.sangari, reshad.hosseini, araabi\}@ut.ac.ir}
}

\renewcommand{\shorttitle}{\textit{arXiv} Template}

\hypersetup{
	pdftitle={Enhancing Interpretability of Sparse Latent Representations with Class Information},
pdfsubject={q-bio.NC, q-bio.QM},
pdfauthor={David S.~Hippocampus, Elias D.~Striatum},
pdfkeywords={First keyword, Second keyword, More},
}

\begin{document}
\renewcommand{\undertitle}{}
\date{}
\maketitle

\begin{abstract}
\setlength{\parindent}{15pt}
\setlength{\parskip}{0pt}
Variational Autoencoders (VAEs) are powerful generative models for learning latent representations. Standard VAEs generate dispersed and unstructured latent spaces by utilizing all dimensions, which limits their interpretability, especially in high-dimensional spaces. To address this challenge, Variational Sparse Coding (VSC) introduces a spike-and-slab prior distribution, resulting in sparse latent representations for each input. These sparse representations, characterized by a limited number of active dimensions, are inherently more interpretable. Despite this advantage, VSC falls short in providing structured interpretations across samples within the same class. Intuitively, samples from the same class are expected to share similar attributes while allowing for variations in those attributes. This expectation should manifest as consistent patterns of active dimensions in their latent representations, but VSC does not enforce such consistency.

In this paper, we propose a novel approach to enhance the latent space interpretability by ensuring that the active dimensions in the latent space are consistent across samples within the same class. To achieve this, we introduce a new loss function that encourages samples from the same class to share similar active dimensions. This alignment creates a more structured and interpretable latent space, where each shared dimension corresponds to a high-level concept, or "factor." Unlike existing disentanglement-based methods that primarily focus on global factors shared across all classes, our method captures both global and class-specific factors, thereby enhancing the utility and interpretability of latent representations. Furthermore, we experimentally demonstrate that classes within the same category (e.g., boots, sandals, and sneakers under the category "shoes") share more common active dimensions, providing deeper insights into the stronger latent similarities among classes within the same category compared to those across different categories. The code is available at \url{https://github.com/farshadsangari/interpretable-latents}


\end{abstract}

\section{Introduction}
\setlength{\parindent}{15pt}
\setlength{\parskip}{0pt}
Artificial Intelligence (AI) systems have revolutionized numerous domains, achieving remarkable success across a wide range of applications. However, as AI models grow increasingly complex, their decision-making processes often become opaque, leading to challenges in trust, accountability, and usability. This phenomenon, commonly referred to as the "black-box problem," has raised significant concerns about the transparency and ethical deployment of AI systems. Interpretability, the ability to understand and explain AI models and internal representations, has thus emerged as a critical research focus, bridging the gap between high performance and practical usability, particularly in high-stakes domains such as healthcare, finance, and autonomous systems.

Interpretability in AI can be broadly categorized into two principal approaches: input space interpretability and latent space interpretability. Input space interpretability focuses on analyzing the relationship between input features and model predictions. Techniques like Grad-CAM (\cite{selvaraju2017grad}), LIME (\cite{ribeiro2016should}), and SHAP (\cite{lundberg2017unified}) exemplify this approach by providing visual or quantitative insights into which parts of the input data influence the model's decisions. For instance, Grad-CAM highlights regions in an image that are most relevant to the model's classification output, offering a direct and visually intuitive explanation. While these methods are valuable for understanding specific predictions, they are inherently tied to the complexity of the input space, where individual features often lack intrinsic interpretability. This limitation makes it challenging to extract high-level, conceptual insights that generalize across samples.

Latent space interpretability, on the other hand, focuses on the internal representations learned by AI models. These latent representations offer a compressed and abstract view of the data, where dimensions can correspond to meaningful, high-level concepts such as object shape, texture, or orientation. For instance, in a model trained on the MNIST dataset (\cite{1571417126193283840}), one latent dimension might encode the thickness of digits, while another captures their rotation. Unlike input space methods, latent space interpretability enables the discovery of generalizable insights, as the learned representations are consistent across samples. Furthermore, the compactness of latent spaces reduces redundancy and noise, enhancing their interpretability and utility in downstream tasks such as classification and generation. Given these advantages, this work focuses on interpretability within the latent space, with an emphasis on generative models.

Generative models, such as VAEs (\cite{kingma2013auto}), GANs (\cite{goodfellow2020generative}), and Diffusion Models (\cite{ho2020denoising}), pose unique interpretability challenges. Some of These models aim to learn latent representations that capture the underlying structure of data, enabling applications such as image synthesis, style transfer, and data augmentation. Understanding the relationship between latent variables and data attributes is crucial for interpreting and controlling the outputs of these models. For example, in a VAE trained on facial images, identifying a latent dimension corresponding to pose allows users to manipulate the angle of generated faces, demonstrating the practical utility of interpretability in generative settings.

Two prominent approaches to latent space interpretability for Generative models are Disentangled Representation Learning and Sparse Coding. Disentanglement based methods (\cite{mathieu2019disentangling, burgess2018understanding, meo2023tc, chen2018isolating, higgins2017beta, kim2018disentangling, adel2018discovering, chen2016infogan, voynov2020unsupervised, ren2021learning, lin2020infogan, sankar2021glowin, esser2020disentangling, yang2301disdiff}) aims to structure the latent space such that each dimension corresponds to a distinct and interpretable factor of variation in the data. This approach ensures that modifying one latent variable affects only a specific attribute while leaving others unchanged. Models like \(\beta\)-VAE, FactorVAE, and InfoGAN encourage disentanglement through regularization techniques or information-theoretic constraints. For instance, \(\beta\)-VAE introduces a penalty term to enforce independence among latent variables, promoting a clear mapping between dimensions and data attributes. Despite their success, these methods often focus on global factors shared across all samples, limiting their ability to capture class-specific attributes.

Sparse coding emphasizes efficiency by representing data as a linear combination of a small set of basis vectors, ensuring that most coefficients are zero or near-zero. This sparsity highlights the most salient features while suppressing redundant information, making it easier to identify meaningful patterns in data. Sparse coding has been widely applied in signal processing, image recognition, and feature extraction due to its interpretability and computational efficiency. However, traditional sparse coding techniques are deterministic and struggle with complex data distributions, necessitating probabilistic extensions like VSC (\cite{tonolini2019variational}).

\begin{figure}[!htbp]
	\centering
	\includegraphics[width=0.6\textwidth]{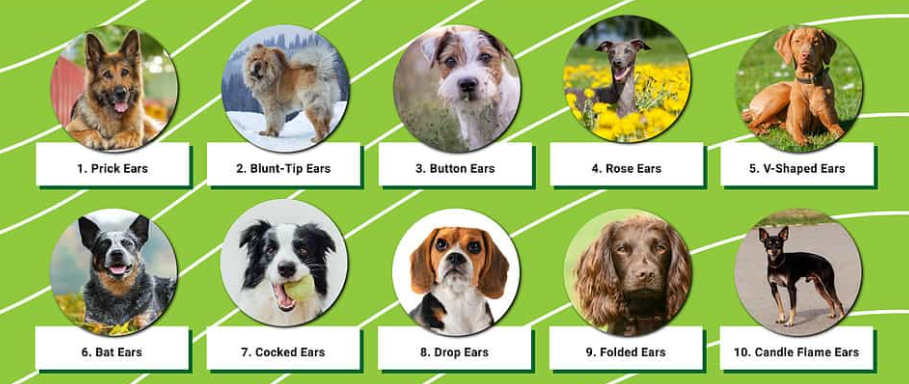}
	\caption{Examples of different types of dog ears categorized by shape and structure. This illustrates that within a class (dogs), common features (such as ears) exist, but their specific characteristics (like shape and size) can vary across samples(\cite{dogears2025})}
	\label{fig:other:dogs-ear}
\end{figure}

VSC enhances latent space interpretability by ensuring that each input activates only a subset of latent dimensions, effectively filtering out irrelevant noise and focusing on essential features. This aligns with the Redundancy Reduction Hypothesis (\cite{barlow1961possible}), which suggests that neural systems minimize redundant information to create more efficient representations. By enforcing sparsity, VSC improves efficiency, which in turn enhances interpretability by reducing latent space clutter and making feature encoding more distinct. Unlike standard VAEs, which utilize all available latent dimensions, VSC selects a structured and compact representation, improving robustness and efficiency. However, while VSC provides a clearer latent structure for individual samples, it does not enforce consistency across samples within the same class. Ideally, data points belonging to the same class should share common latent attributes while allowing for variations in those attributes. For example, all dogs have ears, but the value of this feature, such as the shape and size of the ears, may vary (see Figure \ref{fig:other:dogs-ear}). Since VSC applies sparsity at the individual sample level, it does not guarantee that similar samples activate the same latent dimensions, limiting its interpretability when analyzing class-wide patterns.

Building on the strengths of VSC, we propose a novel approach to enhance latent space interpretability by aligning active latent dimensions across samples within the same class. This alignment ensures that shared attributes are consistently encoded in specific latent dimensions, representing high-level concepts that generalize across class samples. To achieve this, we introduce a new loss function that penalizes discrepancies in active dimensions for class-aligned samples. By enforcing this structure, our method balances global disentanglement and class-specific interpretability, enabling nuanced control and understanding of the latent space.

Experimentally, we demonstrate that our method captures both global and class-specific features, resulting in a more interpretable latent space. For example, we show that in a dataset of footwear (e.g., boots, sandals, sneakers), classes within the same category share common active dimensions, highlighting latent similarities that extend beyond individual samples. This dual-level interpretability offers deeper insights into the structure of data and enhances the practical utility of generative models.

In the following sections, we provide a detailed exploration of our methodology, experimental validation, and the advantages of our proposed framework over existing approaches. By addressing the limitations of current methods, our work advances the state of the art in latent space modeling, contributing to the development of more interpretable, transparent, and reliable AI systems.

\section{Background}
\setlength{\parindent}{15pt}
\setlength{\parskip}{0pt}
\subsection{Disentangled representation learning}
Disentangled representation learning focuses on uncovering latent variables that correspond to distinct, interpretable factors of variation within data. The goal is to separate these factors in such a way that altering one variable affects only one particular aspect of the input, while leaving others unchanged. This makes the representation highly interpretable, as each latent dimension corresponds to a human-understandable characteristic. For example, in a dataset of faces, one latent variable might represent pose, while another captures facial expression. Such disentanglement is crucial for improving the transparency and control of machine learning models, particularly in tasks requiring generative models.

The success of disentangled representations lies in their ability to enhance interpretability and provide a clearer understanding of how generative processes work. By isolating different factors of variation, these representations make it easier to manipulate specific attributes of data during synthesis, analyze model outputs, and transfer learned features across tasks or domains. The independent and structured nature of disentangled representations also facilitates model debugging, making it simpler to track which latent dimensions are responsible for particular outputs.

Several generative approaches, including VAEs (\cite{mathieu2019disentangling, burgess2018understanding, meo2023tc, chen2018isolating, higgins2017beta, kim2018disentangling}), GANs (\cite{adel2018discovering, chen2016infogan, voynov2020unsupervised, ren2021learning, lin2020infogan}), flow-based models (\cite{sankar2021glowin, esser2020disentangling}), and diffusion models (\cite{yang2301disdiff}), tackle the challenge of disentanglement. Among these, VAEs are particularly notable for their structured approach to learning disentangled representations.

VAEs (\cite{kingma2013auto}) are a class of generative models designed for efficient unsupervised learning of latent representations by maximizing a lower bound to the marginal likelihood \( p(x) = \prod p(x_j) \). This optimization is performed with respect to two sets of parameters: the decoding parameters \( \theta \) for the conditional density of input given latent \( p_\theta(x|z) \) and the encoding parameters \( \phi \) of a recognition model \( q_\phi(z|x) \) which is an approximation to the posterior density $p(z|x)$. Commonly, the latent variable \( z \) is supposed to have a gaussian density with zero mean and identity covariance, i.e., \( p(z) = \mathcal{N}(z; 0, I) \). The lower bound to the marginal likelihood is called the Evidence Lower Bound (ELBO), and is expressed as:

\[
\log p_\theta(x) \geq \mathbb{E}_{q_\phi(z|x)}[\log p_\theta(x|z)] - D_{\mathrm{KL}}(q_\phi(z|x) \parallel p(z)),
\]
wherein $\mathbb{E}_q(.)$ is the expectation over probability density $q$ and $D_{KL}()$ is the Kullback-Leibler divergence. The ELBO consists of two terms:
\begin{itemize}
    \item \textbf{Reconstruction term} \( \mathbb{E}_{q_\phi(z|x)}[\log p_\theta(x|z)] \): This measures how well the model reconstructs the data, encouraging the decoder to generate samples that resemble the input data.
    \item \textbf{KL divergence term} \( D_{\mathrm{KL}}(q_\phi(z|x) \parallel p(z)) \): This term acts as a regularizer, ensuring that the learned latent density \( q_\phi(z|x) \) stays close to the prior \( p(z) \).
\end{itemize}

\subsection{Variational Sparse coding}

The paper titled "Variational Sparse Coding" (\cite{tonolini2019variational}) introduces a specific prior distribution that improves the interpretability and efficiency of VAEs by integrating sparse coding. Traditional VAEs often generate dense latent spaces, leading to encode a sample in all latent space. To address this, the authors propose VSC, which introduces a Spike and Slab prior to induce sparsity in the latent space, ensuring that only a subset of dimensions is active for each data point.

\begin{figure}[!htbp]
	\centering
	\includegraphics[width=0.8\textwidth]{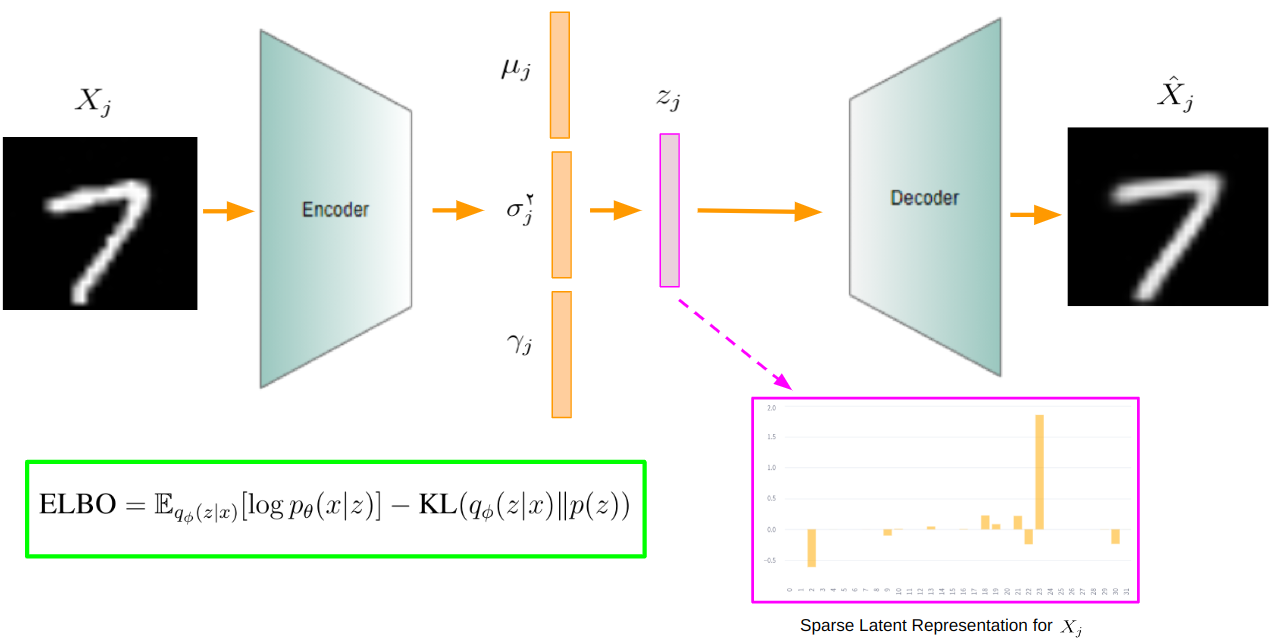}
	\caption{VSC architecture. The encoder $q_{\phi}(z|x)$ outputs the mean, standard deviation, and sparsity parameters $\gamma$, encouraging sparse latent representations. The decoder $p_{\theta}(x|z)$ reconstructs the input. Compared to standard VAEs, VSC promotes interpretability by activating only a subset of latent dimensions.}
	\label{fig:our:vsc}
\end{figure}

The Spike and Slab prior is defined as:
\begin{equation}
	\begin{aligned}
p(z) = \prod_{i=1}^{d} \left( \alpha \mathcal{N}(z_i; 0, 1) + (1 - \alpha) \delta(z_i) \right),
	\end{aligned}
\end{equation}
where $\delta(.)$ is the dirac function and \( \alpha \) controls the probability that a latent dimension is active. The KL divergence term in the ELBO in the term responsible for creating a sparse representation in the latent space:

\begin{equation}
	\begin{aligned}
		-\text{KL}(Q || P) = \sum_{i=1}^{d} \bigg[ &\gamma_i \left( \frac{1}{2} \left( 1 + \log \sigma_i^2 - \mu_i^2 - \sigma_i^2 \right) \right) \\
		&+ (1 - \gamma_i) \log \left( \frac{1 - \alpha}{1 - \gamma_i} \right) + \gamma_i \log \left( \frac{\alpha}{\gamma_i} \right) \bigg],
	\end{aligned}
\end{equation}
where $\gamma_i$ represents the probability of the jth latent variable is active for the input $x$. The final loss function for the input data $x_j,\ j=1,\hdots,n$ can be written as the Equation \ref{eq:3:vsc_total_loss_func}. 

\begin{equation}
	\begin{aligned}
		\mathcal{L}(\theta, \phi; x_{j}) \simeq & \sum_{i=1}^{d} \Bigg[ \frac{\gamma_{i,j}}{2} \left( 1 + \log(\sigma_{i,j}^{2}) - \mu_{i,j}^{2} - \sigma_{i,j}^{2} \right) \\
		& + (1 - \gamma_{i,j}) \log \left( \frac{1-\alpha}{1-\gamma_{i,j}} \right) + \gamma_{i,j} \log \left( \frac{\alpha}{\gamma_{i,j}} \right) \Bigg] 
		 + \frac{1}{L} \sum_{l=1}^{L} \log p_{\theta}(x_{j}|z_{l,j}).
	\end{aligned}
\label{eq:3:vsc_total_loss_func}
\end{equation}

Compared to traditional VAEs, \textit{VSC} provides several key benefits:
\begin{itemize}
    \item \textbf{Improved Interpretability}: Sparse representations enable easier identification of which latent dimensions control specific features, making the model more interpretable.
    \item \textbf{Efficiency}: VSC uses fewer active dimensions to represent data, resulting in more efficient encoding, which is beneficial for tasks like classification.
    \item \textbf{Robustness}: VSC remains stable even as the number of latent dimensions increases, as it selectively activates only the most important parts of the latent representation while ignoring noise. This selective encoding prevents overfitting to irrelevant variations and enhances the model's resilience compared to standard VAEs, which tend to degrade in performance as the latent space grows.
\end{itemize}

\section{Methodology}
\setlength{\parindent}{15pt}
\setlength{\parskip}{0pt}
In the previous section, we see VSC architecture and their main idea. In VSC, the spike variable, which indicates whether each dimension is active or not, is drawn from a Bernoulli distribution with a success probability of gamma. Therefore, for each input, the gamma vector, which represents the probability of each latent space dimension being active or inactive, can be considered. The VSC paper (\cite{tonolini2019variational}) argued that this approach leads to a more sparse and, as a result, more interpretable representation for each input.

\begin{figure}[!htbp]
	\centering
	\includegraphics[width=0.6\textwidth]{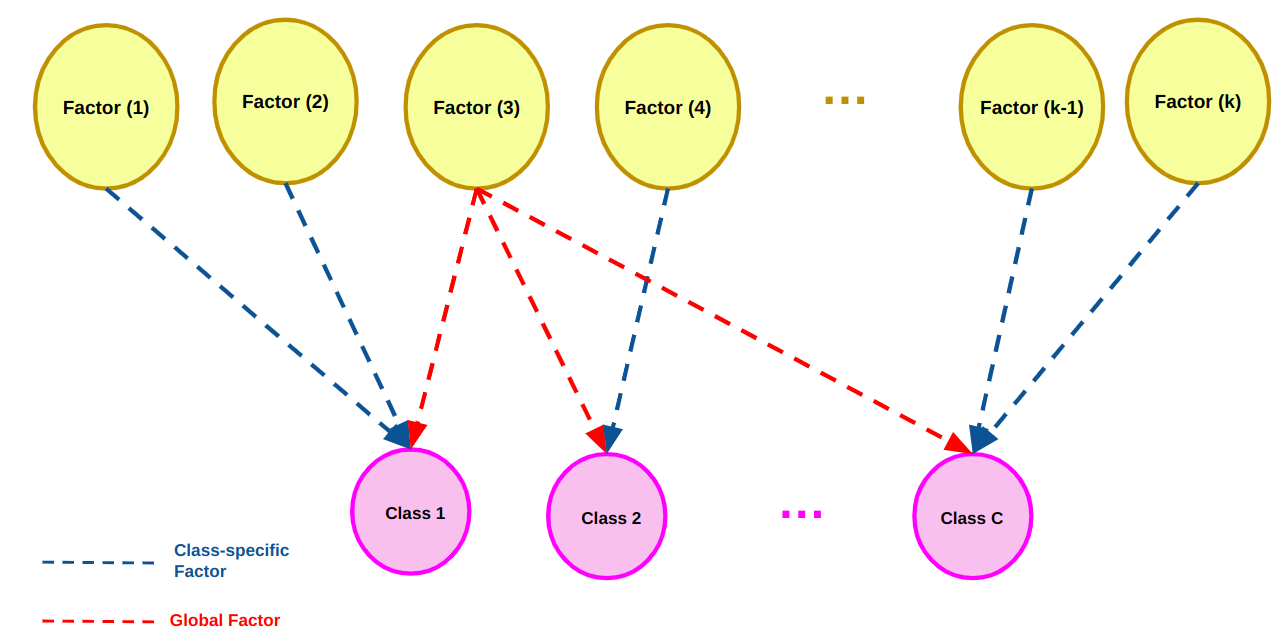}
	\caption{Illustration of global and class-specific factors. Global factors are shared across all classes, while class-specific factors vary between classes. Each sample can be represented by a combination of global and class-specific active latent dimensions.}
	\label{fig:our:idea_graph}
\end{figure}

In the real world, objects within a class share common features; for example, all dogs have ears, but the value of this feature, such as the shape and size of the ears, may vary (see Figure \ref{fig:other:dogs-ear}). In generative models that create a compact representation for input data, each of the features in the latent space can represent a specific visual feature. For each class of data, it is expected that the shared features play a role, and the difference between the data of each class arises from the differences in the values of these shared features. In the VSC method, we see that each input uses a portion of the latent space's features. Our main idea is that these active dimensions should largely be shared for the data of each class. In fact, within a dataset, classes can not only share common features but also possess their own unique characteristics which is illustrated in the figure \ref{fig:our:idea_graph}. 

To achieve this goal, we build upon the VSC model, which introduces a spike variable to enforce sparsity in the latent space. Each dimension's spike variable follows a Bernoulli distribution parameterized by gamma. To align the spike distribution across each dimension of the latent space for data in a class, the Kullback-Leibler divergence can be used. Additionally, due to symmetry and the existence of an upper bound, using the Jensen-Shannon distance (\cite{menendez1997jensen}) is also an appropriate option in order to avoid instability. This loss term is compatible with other terms in the VSC model and maintains coherence in optimization. In fact, considering \ref{eq:3:vsc_total_loss_func}, which is the loss function of the VSC method, we observe that all terms in this loss function are based on Kullback-Leibler divergence.

\begin{figure}[!htbp]
	\centering
	\includegraphics[width=0.72\textwidth]{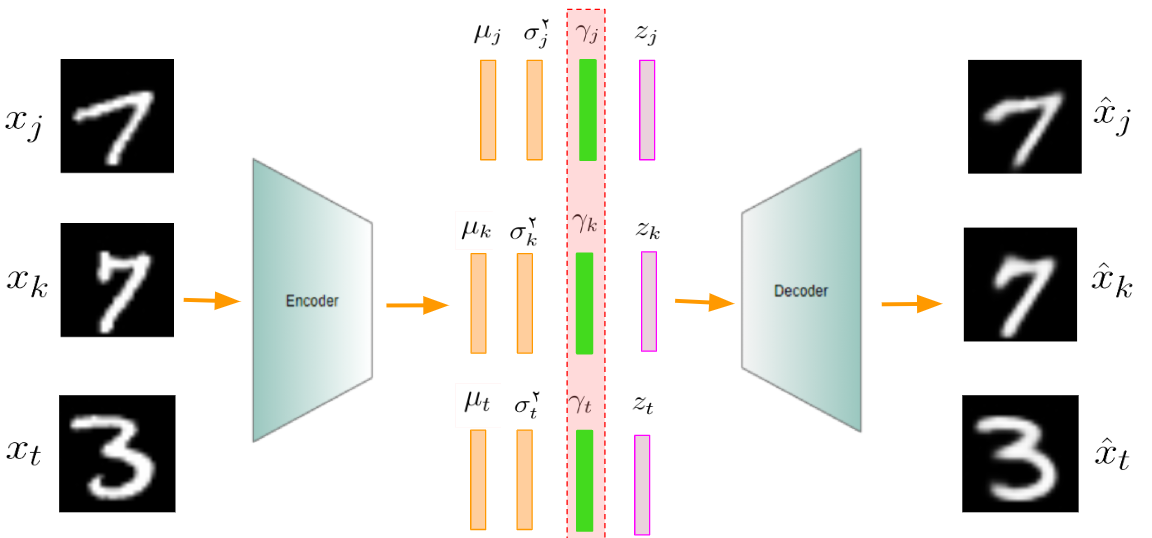}
	\caption{Proposed method: aligning active latent dimensions for samples within the same class. For each pair of samples, the Jensen-Shannon distance is computed between their spike probability vectors $\gamma$, encouraging similar activation patterns among class members.}
	\label{fig:our:idea_overview}
\end{figure}

\begin{figure}[!htbp]
	\centering
	\includegraphics[width=0.8\textwidth]{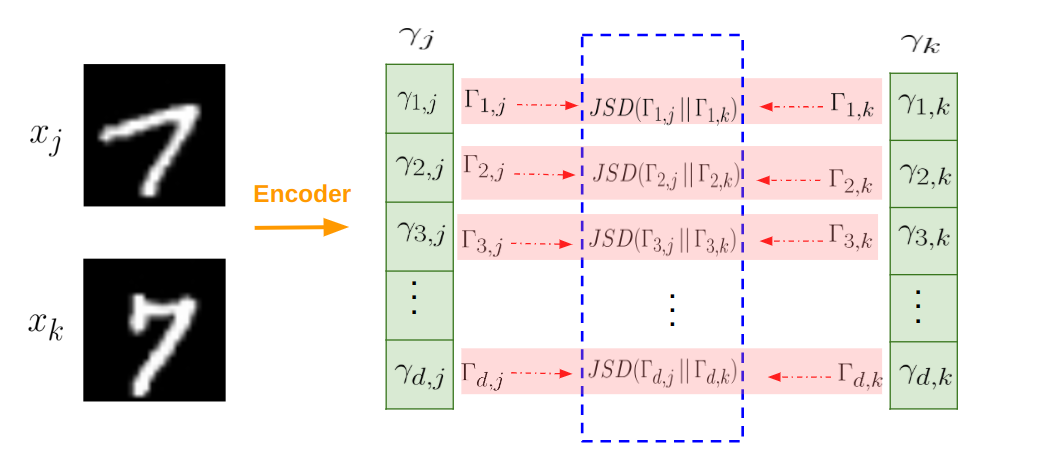}
	\caption{Calculation of the Jensen-Shannon distance for two samples within the same class. The spike probabilities $\gamma$ for each dimension are compared independently under the assumption of latent dimension independence, and their average is used as the similarity measure.}
	\label{fig:our:idea_2}
\end{figure}

According to Figure \ref{fig:our:idea_overview}, suppose the vectors \(\gamma_{j} = [\gamma_{1,j}, \gamma_{2,j}, \ldots, \gamma_{d,j}]\) and \(\gamma_{k} = [\gamma_{1,k}, \gamma_{2,k}, \ldots, \gamma_{d,k}]\) represent the probabilities of the latent space dimensions being active for two data points \(j\) and \(k\) in the common class \(c\). Each dimension of these gamma vectors is the probability parameter of the spike variable, which follows a Bernoulli distribution(\(\Gamma_{i,k}\)) for that dimension of the input data in the latent space. For instance, \(\gamma_{i,k}\) is the probability that the \(i\)-th dimension of the \(k\)-th input data is active. Given the assumption of independence between the dimensions in the latent space, for any two data points within a class, the Jensen-Shannon distance can be calculated separately for each dimension, and the summation can be taken, as shown in Figure \ref{fig:our:idea_2}. Thus, for two data points \(j\) and \(k\) in a class, we have:
\begin{equation}
	\text{JSD}(\Gamma_{j} \parallel \Gamma_{k}) = \sum_{i=1}^{d} \text{JSD}(\Gamma_{i,j} \parallel \Gamma_{i,k}) = \sum_{i=1}^{d} \left( \frac{1}{2} D_{KL}(\Gamma_{i,j} \parallel M_i) + \frac{1}{2} D_{KL}(\Gamma_{i,k} \parallel M_i) \right),
	\label{eq:4:js_ours_2}
\end{equation}
where \(d\) represents the number of latent space dimensions and \(M_i\) is the mean of the two distributions. Given that the spike variable follows a Bernoulli distribution, we can simplify this term and consider a closed form for it. In this case, we arrive at Equation \ref{eq:4:js_ours_3}.

\begin{equation}
	\begin{aligned}
		JSD(\Gamma_{j} \parallel \Gamma_{k}) = \sum_{i=1}^{d} \frac{1}{2} & \left[ \gamma_{i,j} \log \left( \frac{2 \gamma_{i,j}}{\gamma_{i,j} + \gamma_{i,k}} \right) + (1 - \gamma_{i,j}) \log \left( \frac{2(1 - \gamma_{i,j})}{2 - \gamma_{i,j} - \gamma_{i,k}} \right) \right] \\
		& + \frac{1}{2} \left[ \gamma_{i,k} \log \left( \frac{2 \gamma_{i,k}}{\gamma_{i,j} + \gamma_{i,k}} \right) + (1 - \gamma_{i,k}) \log \left( \frac{2(1 - \gamma_{i,k})}{2 - \gamma_{i,j} - \gamma_{i,k}} \right) \right].
	\end{aligned}
\label{eq:4:js_ours_3}
\end{equation}

\begin{figure}[!htbp]
	\centering
	\includegraphics[width=0.7\textwidth]{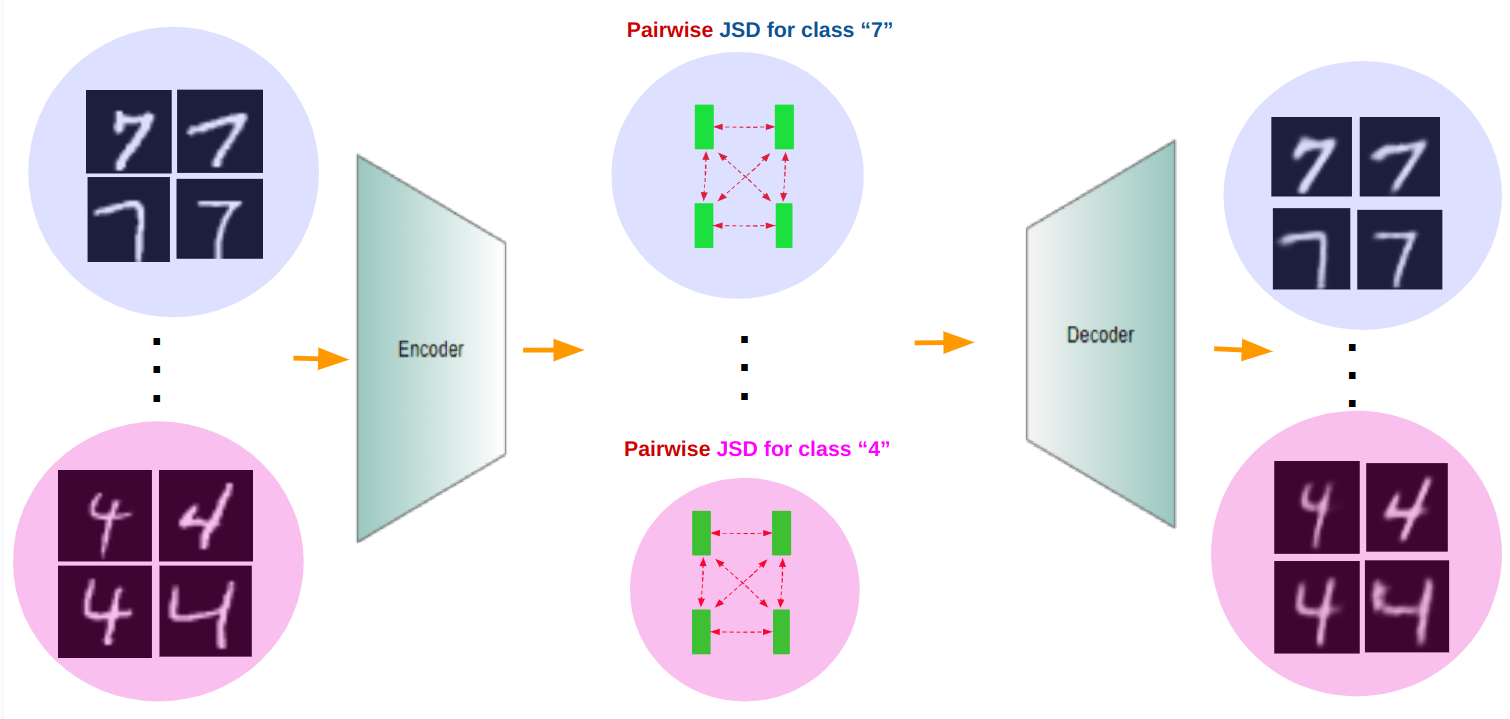}
	\caption{Computation of the proposed Jensen-Shannon loss term across each class. For each class, the pairwise Jensen-Shannon distances between samples are averaged to encourage alignment of active latent dimensions among all class members.}
	\label{fig:our:idea_classwise}
\end{figure}

If we calculate this equation for each category, average it for the data in each class, we arrive at Equation \ref{eq:4:loss_js_1}. As shown in Figure \ref{fig:our:idea_classwise}, for each category of input data, for example, the classes of digits 9 and 4, the proposed Jensen-Shannon distance term can be computed and averaged for their gamma vectors in the latent space.

\begin{equation}
	\text{L}_{\text{JSD}} = \frac{1}{|C|} \sum_{c \in \mathcal{C}} \frac{1}{N_c} \sum_{\substack{(k,j) \in c \\ k \neq j}} \text{JSD}(\Gamma_{j} \parallel \Gamma_{k}),
	\label{eq:4:loss_js_1}	
\end{equation}
where \(C\) and \(N_c\) refer to the classes and number of data pairs in class \(c\), respectively.

By combining the VSC loss function and Equation \ref{eq:4:loss_js_1}, we arrive at the final loss function Equation \ref{eq:4:loss_js_2}.
\begin{equation}
    \begin{aligned}
        L_{\text{total}} = L_{\text{VSC}} + \lambda L_{\text{JSD}} 
        = & - \sum_{i=1}^{d} \Bigg[ \frac{\gamma_{i,j}}{2} \left( 1 + \log(\sigma_{i,j}^{2}) 
        - \mu_{i,j}^{2} - \sigma_{i,j}^{2} \right) \\
        & + (1 - \gamma_{i,j}) \log \left( \frac{1-\alpha}{1-\gamma_{i,j}} \right) 
        + \gamma_{i,j} \log \left( \frac{\alpha}{\gamma_{i,j}} \right) \Bigg] \\
        & - \frac{1}{L} \sum_{l=1}^{L} \log p_{\theta}(x_{j}|z_{l,j}) \\
        & + \frac{\lambda}{|C|} \sum_{c \in \mathcal{C}} \frac{1}{N_c} 
        \sum_{\substack{(k,t) \in c \\ k \neq t}} \text{JSD}(\Gamma_{k} \parallel \Gamma_{t}),
        \label{eq:4:loss_js_2}
    \end{aligned}
\end{equation}
where \(\lambda\) is the hyperparameter. The VSC loss function leads to sparsification of the latent space. On the other hand, the proposed term helps to allign the active dimensions for each class of data. As a result, the combination of these two terms ensures that the same latent dimensions are used as much as possible for the data in each class. Essentially, this means that the spike variables of samples within each class should have the same values as much as possible.
The algorithm for the proposed method is shown in Algorithm \ref{algo:ours:1}.

\begin{algorithm}[H] 
	\caption{Algorithm of proposed method} 
	\label{algo:ours:1}
	\begin{algorithmic}[1]
		\REQUIRE Initialize parameters $\theta$
		\REQUIRE Training set $\mathcal{D} = \{(\mathbf{x}_i, y_i)\}_{i=1}^N$
		
		\FOR{each epoch}
		\FOR{each batch}
		\STATE Initialize $L_{\text{JSD}} \leftarrow 0$
		\STATE Initialize $L_{\text{VSC}} \leftarrow 0$
		
		\FOR{each class $c$ in batch}
		\STATE Initialize $JS_{\text{class}} \leftarrow 0$
		\STATE Count pairs $N_{\text{pairs}} \leftarrow 0$
		
		\FOR{all pairs $(x_j, x_k)$ in class $c$ where $j \neq k$}
		\STATE Calculate $\gamma_j$ and $\gamma_k$ using the Encoder.
		\STATE Compute $JSD(\Gamma_j \parallel \Gamma_k)$ as defined in Equation \ref{eq:4:js_ours_3}
		\STATE $JS_{\text{class}} \leftarrow JS_{\text{class}} + JSD(\Gamma_j \parallel \Gamma_k)$
		\STATE $N_{\text{pairs}} \leftarrow N_{\text{pairs}} + 1$ 
		\ENDFOR
		
		\STATE $JS_{\text{class}} \leftarrow \frac{JS_{\text{class}}}{N_{\text{pairs}}}$
		\STATE $L_{\text{JSD}} \leftarrow L_{\text{JSD}} + JS_{\text{class}}$
		\ENDFOR
			
		\STATE $L_{\text{JSD}} \leftarrow \frac{L_{\text{JSD}}}{\text{C}}$
		
		\STATE $L_{\text{VSC}} \leftarrow \text{Calculate Average } L_{\text{VSC}} \text{ for the batch (Equation \ref{eq:3:vsc_total_loss_func})}$
		
		\STATE $L_{\text{total}} \leftarrow L_{\text{VSC}} + \lambda L_{\text{JSD}}$ 
		
		\STATE Calculate gradients to minimize loss function
		\STATE Update parameters $\theta$ based on the gradients
		\ENDFOR
		\ENDFOR
	\end{algorithmic}
\end{algorithm}

\section{Experimental Results}
\setlength{\parindent}{15pt}
\setlength{\parskip}{0pt}

We evaluate our method using two widely recognized image datasets for benchmarking performance: the MNIST dataset (\cite{1571417126193283840}), which consists of handwritten digits, and the more recent Fashion-MNIST dataset (\cite{xiao2017fashion}), featuring images of various clothing items. Both datasets contain \(28 \times 28\) grayscale images, making them suitable for comparison across different domains.

The MNIST dataset is a commonly used benchmark in disentangled representation learning, as it contains both global and class-specific features. The global features, such as thickness, rotation, and size of the digits, are consistent across all classes. These global factors are crucial for generative models and are typically captured by disentanglement-based methods. Since our method is designed to capture both global and class-specific features, MNIST provides a useful test case to evaluate whether our method can capture the same global factors that disentanglement methods do. In addition to global features, MNIST also contains class-specific features (e.g., the different shapes of digits) that standard disentanglement methods may struggle to capture. This makes MNIST an ideal dataset to evaluate the effectiveness of our method in distinguishing between global and class-specific patterns.

In contrast, Fashion-MNIST does not exhibit a global feature that is consistent across all classes. Due to this, it is less frequently used in traditional disentanglement representation learning methods, which typically focus on global features. However, Fashion-MNIST is a perfect choice for evaluating our method. This dataset consists of several categories of clothing, each with unique visual features. For instance, the "shoes" category includes boots, sneakers, and sandals, which share common features like sole structure and shoe shape. Similarly, the "upper clothing" category contains items like coats, dresses, shirts, t-shirts, and pullovers, which share features such as fabric texture and sleeve structure. Fashion-MNIST allows us to investigate whether our method can capture common latent features within visually similar classes, such as boots and sneakers, which share similar visual patterns, while also differentiating between classes that do not share as many common features. This makes Fashion-MNIST an ideal dataset to demonstrate that classes with similar visual features should activate common latent dimensions in our model.

In the following subsections, we evaluate our model’s performance. First, we analyze feature alignment within each class and compare the results with the VSC model. Second, we quantitatively assess the efficiency of the latent space representation by examining both global and class-specific features, comparing them to those obtained through disentangled representation learning (\cite{chen2016infogan,ren2021learning}). Finally, we qualitatively explore the relationships between sparse feature vectors across different classes.

\subsection{MNIST}
Unlike Fashion-MNIST, the MNIST dataset exhibits clear global factors, as digits share common geometric properties such as rotation, thickness, and vertical scaling. These shared attributes make MNIST particularly suitable for evaluating models that aim to capture both global and class-specific features.

\begin{figure}[!htbp]
	\centering
	\includegraphics[width=0.7\textwidth]{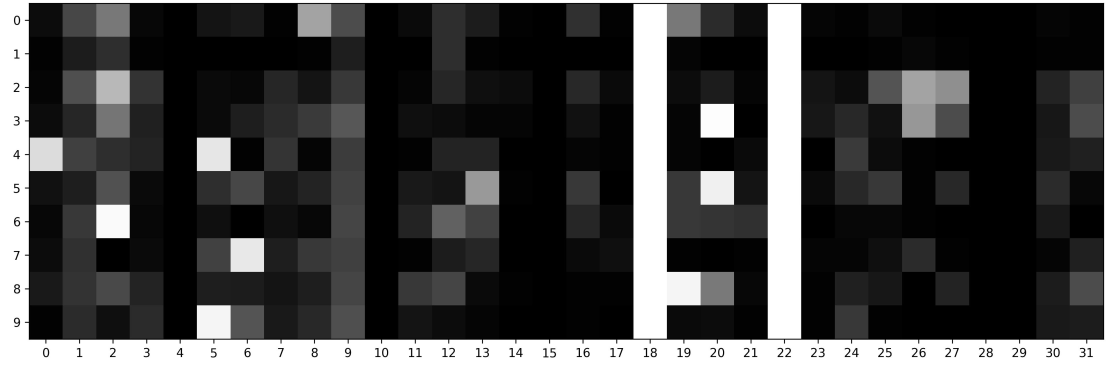}
	\caption{Average gamma probabilities across classes in the MNIST dataset using VSC. The active latent dimensions are not consistently aligned within each class, limiting interpretability.}
	\label{fig:res:mnist_vsc_probs}
\end{figure}

Figure~\ref{fig:res:mnist_vsc_probs} illustrates the active dimensions in the VSC method \citep{tonolini2019variational} across different MNIST classes. The horizontal axis represents latent dimensions, and the vertical axis corresponds to digit classes. Each cell’s value reflects the average activation level of a dimension for a given class. As shown, the active dimensions are not consistently aligned within each class, which limits the interpretability of the VSC model in capturing global or class-specific semantics.

\begin{figure}[!htbp]
	\centering
	\includegraphics[width=0.75\textwidth]{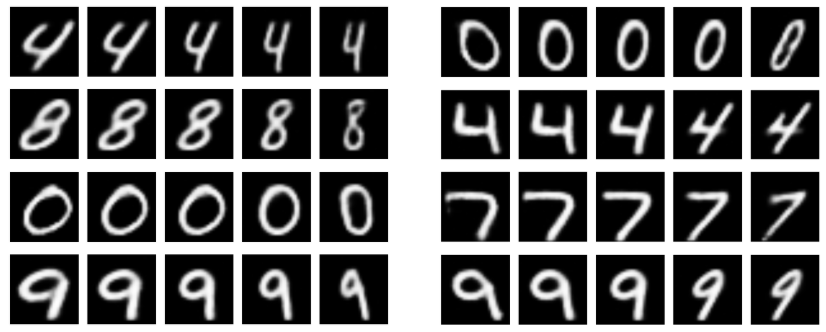}
	\caption{Global features discovered in the MNIST dataset using VSC. Left: The twenty-second latent dimension controls vertical contraction. Right: The eighteenth latent dimension projects digits along the $y = x$ diagonal, indicating global rotation.}
	\label{fig:res:mnist_global_vsc}
\end{figure}

\begin{figure}[!htbp]
	\centering
	\includegraphics[width=0.8\textwidth]{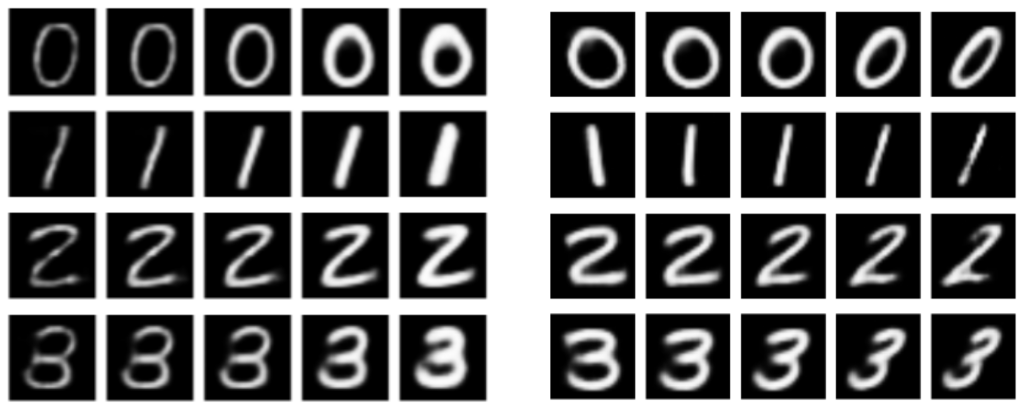}
	\caption{Global features discovered in the MNIST dataset using our proposed method. Left: The second latent dimension controls digit thickness. Right: The eighteenth latent dimension projects digits along the $y=x$ diagonal, indicating global rotation.}
	\label{fig:res:mnist:ours:main_globals}
\end{figure}

Interestingly, dimensions 18 and 22 appear to be active across all classes. This observation suggests that these dimensions may encode global features, such as digit rotation or vertical compression. Latent traversals confirm this: dimension 18 captures rotation, and dimension 22 captures vertical contraction (see Figure~\ref{fig:res:mnist_global_vsc}). However, this method does not capture another important global factor—digit thickness—which we find this dimension in our method (see Figure~\ref{fig:res:mnist:ours:main_globals}, left).

\begin{figure}[!htbp]
	\centering
	\includegraphics[width=0.8\textwidth]{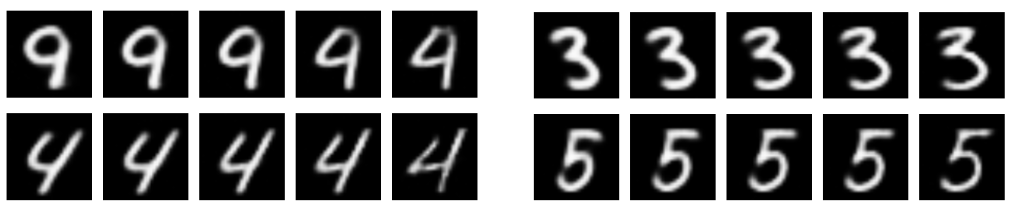}
	\caption{Class-specific features discovered in the MNIST dataset using VSC. Left: The fifth latent dimension may controls the thickness for the digits 4 and 9. Right: The twentieth latent dimension controls the size of the lower circle for digits 3 and 5.}
	\label{fig:res:mnist:vsc_local}
\end{figure}

As shown in Figure~\ref{fig:res:mnist_vsc_probs}, the heatmap for the MNIST classes is not concentrated, meaning that we cannot identify a consistent mask for each class with high gamma probability. To further explore this, we perform latent traversal on two common dimensions, the results of which are shown in Figure~\ref{fig:res:mnist:vsc_local}.

\begin{figure}[!htbp]
	\centering
	\includegraphics[width=0.7\textwidth]{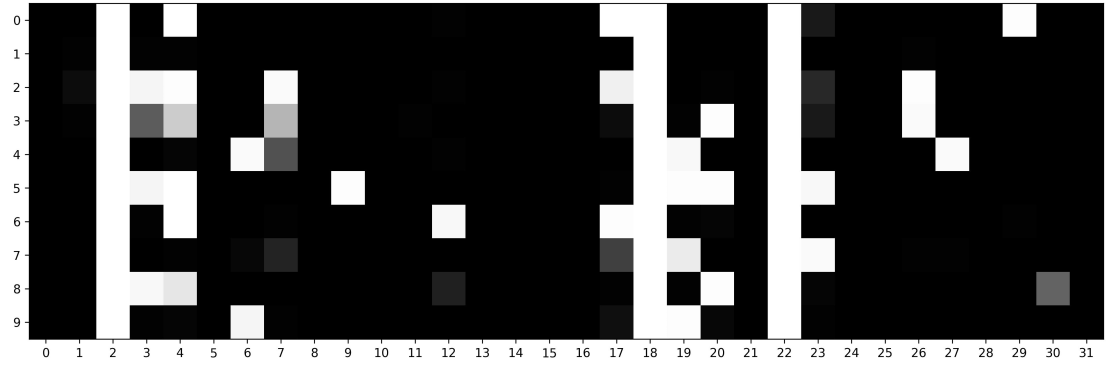}
	\caption{Average gamma probabilities across classes in the MNIST dataset using our proposed method. Active dimensions are well-aligned within each class, improving latent space interpretability.}
	\label{fig:res:mnist_ours_probs}
\end{figure}

We next examine the results of our proposed method. Figure \ref{fig:res:mnist_ours_probs} shows the average gamma probability for each class as determined by our method for the MNIST dataset. As we expected, the active dimensions of the latent space are well aligned for each class. According to this diagram, we observe that some active dimensions are common across all classes. This observation reminds us that these dimensions may encode global features among the classes. These features could be the same ones provided by disentangled representation learning methods. For example, these dimensions may encode the thickness and rotation of the digits.

To validate these observations, we apply the latent traversal technique. We first examine the second dimension, which is common across all classes. To do this, we pass sample data through the encoder to obtain their latent space representation. Then, by altering the second dimension in the latent space, we observe the effect of this change on the decoder's output. According to Figure \ref{fig:res:mnist:ours:main_globals} (left), we see that this dimension encodes the thickness attribute of the digits across different classes. As shown in this figure, this attribute is consistent among all classes. Decreasing the value of this dimension results in thinner digits, while increasing it leads to thicker digits.

Similarly, we investigate the eighteenth dimension, another commonly active feature. According to the results presented in Figure \ref{fig:res:mnist:ours:main_globals} (right), increasing the value of this dimension projects the input image onto the line y=x. This figure indicates that this feature is a global and common attribute among the classes. We provide additional experimental results in Appendix~\ref{appendix:mnist:global}.

\begin{figure}[!htbp]
	\centering
	\includegraphics[width=0.8\textwidth]{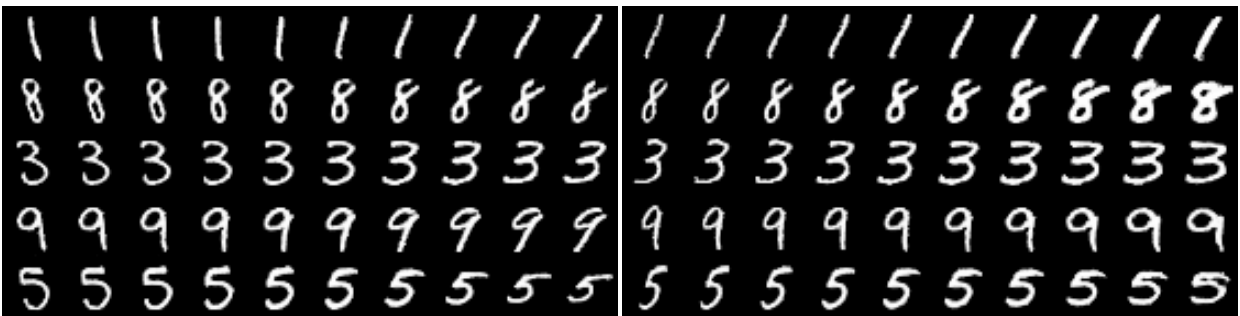}
	\caption{Global disentangled features in MNIST obtained from \cite{chen2016infogan}.}
	\label{fig:res:mnist:infogan-globals}
\end{figure}

\begin{figure}[!htbp]
	\centering
	\includegraphics[width=0.5\textwidth]{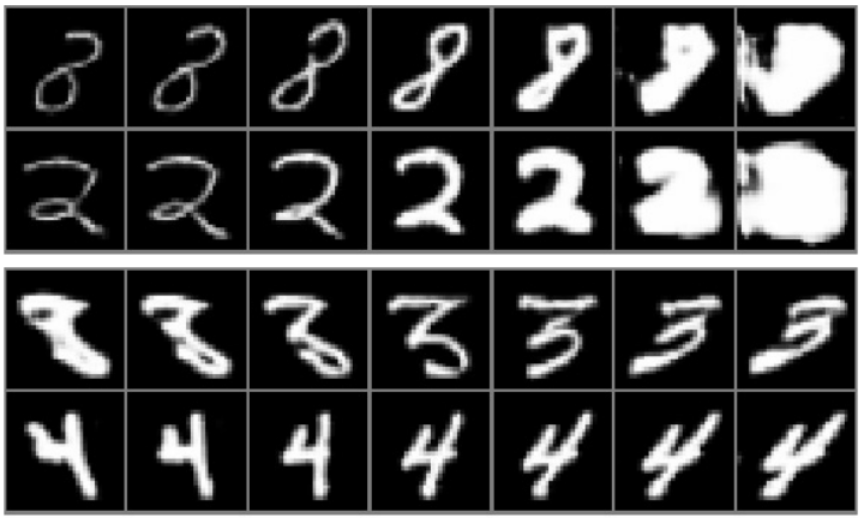}
	\caption{Global disentangled features in MNIST obtained from Ren et al.\ (2021).}
	\label{fig:res:mnist:rengan-globals}
\end{figure}

These findings further reinforce the presence of global attributes consistent across different classes. In fact, all the different classes share common geometric features such as thickness and compactness. These features were also accessible through interpretable representation learning methods(see Figures \ref{fig:res:mnist:infogan-globals} and \ref{fig:res:mnist:rengan-globals})

Beyond global features, our proposed method also captures class-specific interpretations. While some latent dimensions remain active across all classes, others encode unique attributes relevant only to specific classes. Next, we examine and interpret these class-specific dimensions.

\begin{figure}[!htbp]
	\centering
	\includegraphics[width=0.6\textwidth]{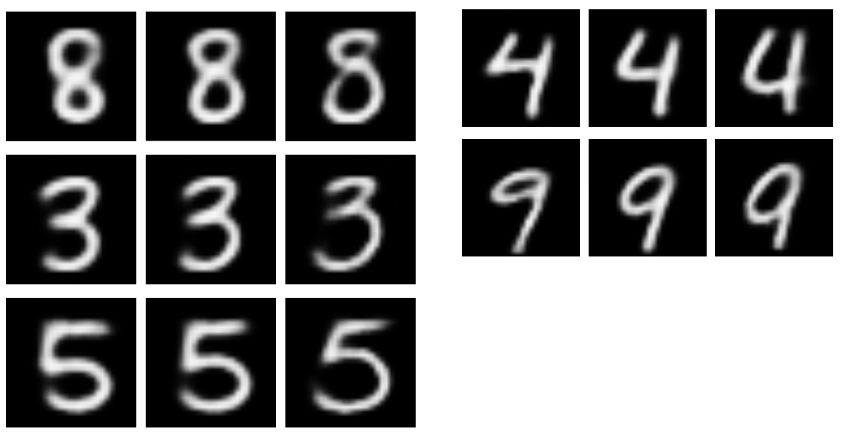}
	\caption{Class-specific features discovered in the MNIST dataset using our proposed method. Left: The 20th dimension controls the lower circle size for digits 3, 5, and 8. Right: The 6th dimension adjusts the intersection position in digits 4 and 9.}
	\label{fig:res:mnist:ours:class-specific-1}
\end{figure}

For instance, Figure \ref{fig:res:mnist_ours_probs} shows that the sixth dimension is commonly active for digits 4 and 9. Using latent traversal, we can determine its meaning. As illustrated in Figure \ref{fig:res:mnist:ours:class-specific-1} (right), this dimension encodes the intersection point between the horizontal and vertical strokes of the digit. Increasing its value shifts this intersection downward. Since other classes lack this structural feature, this interpretation is specific to digits 4 and 9. Similarly, Figure \ref{fig:res:mnist_ours_probs} indicates that the twentieth dimension is commonly active for digits 8, 3, and 5. As seen in Figure \ref{fig:res:mnist:ours:class-specific-1} (left), this dimension encodes the size of the lower circular component in these digits. We provide additional experimental results in Appendix~\ref{appendix:mnist:local}.

To further quantify class-wise latent space similarities, we compare the gamma vectors of different classes using various distance metrics. Specifically, we employ Cosine Distance, Euclidean Distance, and the Pearson Correlation Coefficient. In this section, we report only the Pearson Correlation Coefficient, while results for the other metrics are provided in the Appendix~\ref{appendix:mnist:heatmaps}.

Figure \ref{fig:res:mnist:pearson_coef_main} presents the Pearson Correlation Coefficient between gamma vectors of different classes. The results align with previous findings, confirming class-wise similarities. For example, as previously observed, the gamma vector for digit 9 closely resembles those of digits 1, 4, and 7. This consistency further validates the structured alignment of active latent dimensions within each class.

\begin{figure}[!htbp]
	\centering
	\includegraphics[width=0.5\textwidth]{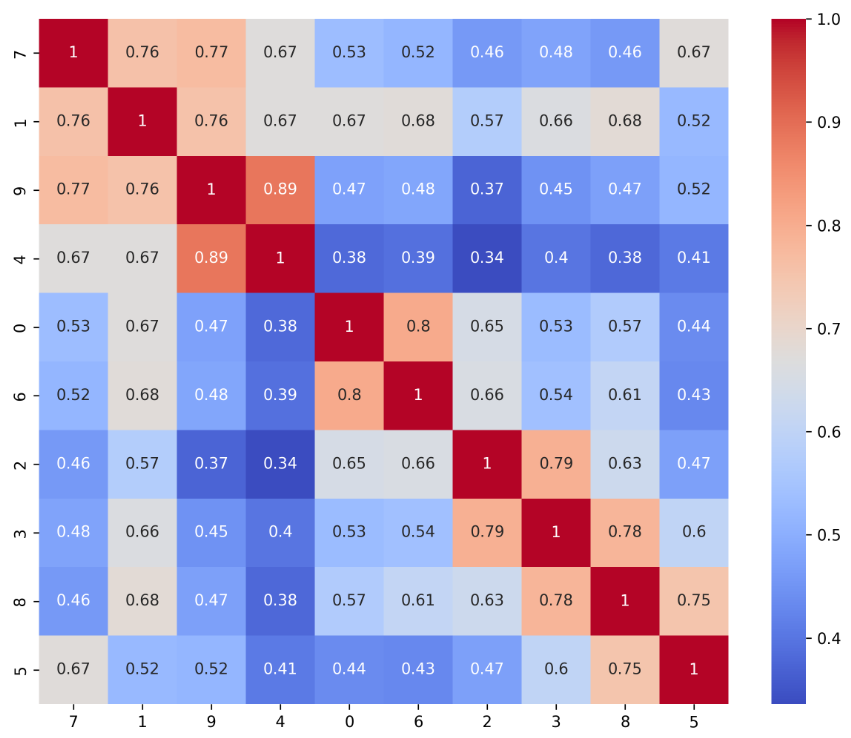}
	\caption{Pearson correlation coefficient between gamma vectors across different MNIST classes using our proposed method. Higher values indicate stronger feature similarity across classes.}
	\label{fig:res:mnist:pearson_coef_main}
\end{figure}

\subsection{Fashion-MNIST}
\begin{figure}[!htbp]
	\centering
	\includegraphics[width=0.7\textwidth]{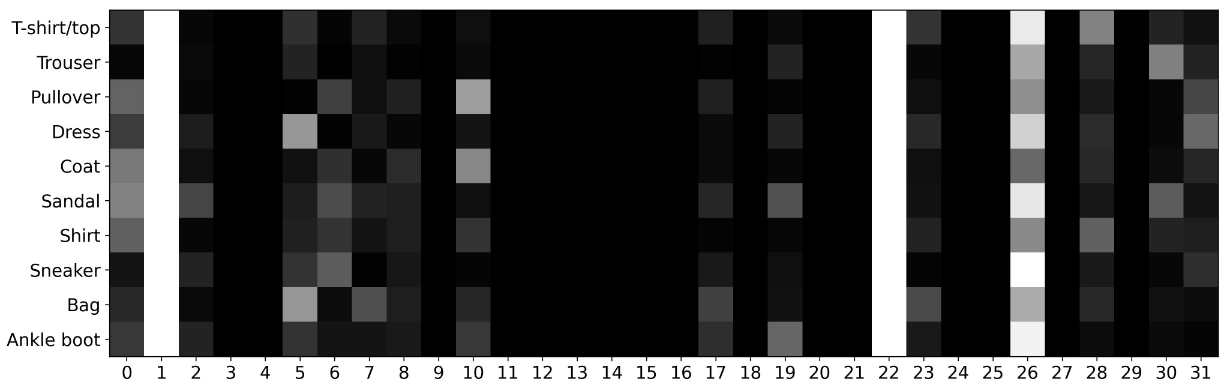}
	\caption{Average gamma probabilities across classes in the Fashion-MNIST dataset using VSC. Active latent dimensions are not consistently aligned within each class.}
	\label{fig:res:fmnist:heatmap_prob}
\end{figure}

Figure \ref{fig:res:fmnist:heatmap_prob} illustrates the activation patterns of latent dimensions in the sparse coding method for Fashion-MNIST classes. As seen in the figure, the active dimensions are not well-aligned within any single class. This misalignment limits the sparse coding method’s ability to provide meaningful interpretations, whether global or class-specific. According to Figure \ref{fig:res:fmnist:heatmap_prob}, the 1st and 22th dimensions remain consistently active. This suggests they may correspond to general features. However, latent traversal reveals that these dimensions do not encode a common, interpretable feature across classes.

\begin{figure}[!htbp]
	\centering
	\includegraphics[width=0.7\textwidth]{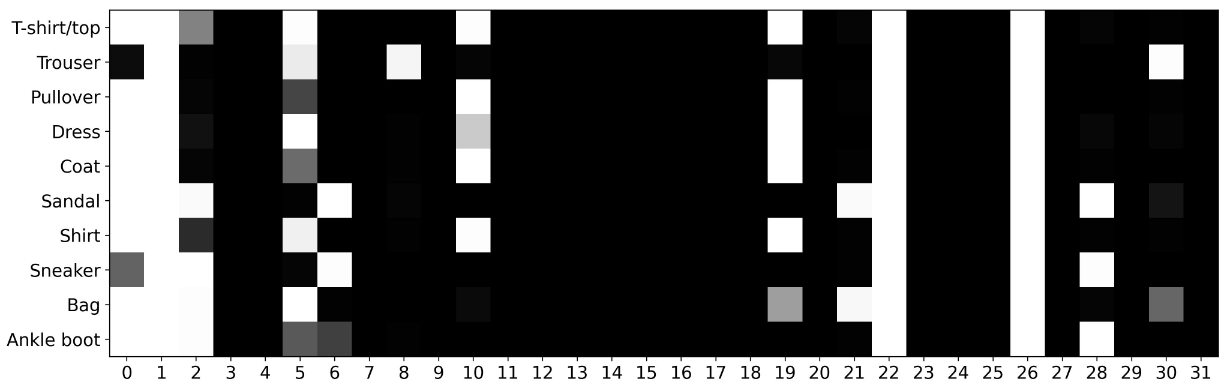}
	\caption{Average gamma probabilities across classes in the Fashion-MNIST dataset using our proposed method. Active dimensions are highly aligned within each class, supporting interpretable latent structures.}
	\label{fig:res:fmnist_ours_probs}
\end{figure}

We now evaluate the results of our proposed method. Figure \ref{fig:res:fmnist_ours_probs} presents the average gamma probability for each class in the Fashion-MNIST dataset. As anticipated, the active latent space dimensions are highly aligned within each class, with certain features being unique to specific classes. In the following section, we will analyze these class-specific features in greater detail.

\begin{figure}[!htbp]
	\centering
	\includegraphics[width=0.7\textwidth]{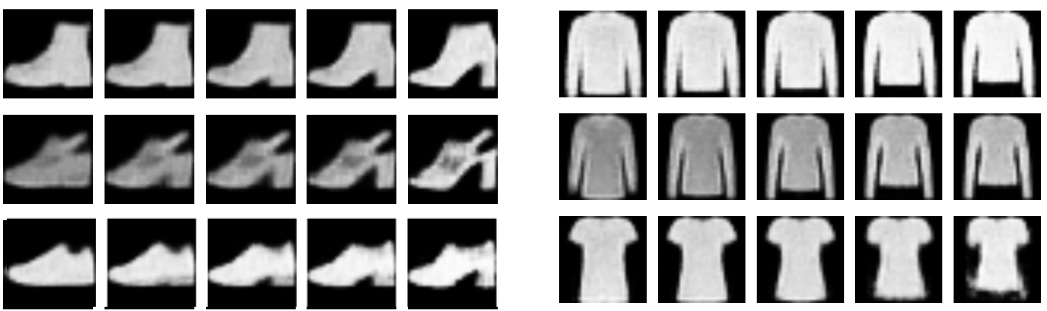}
	\caption{Class-specific interpretations in the Fashion-MNIST dataset using our proposed method. The 10th latent dimension captures the length of upper clothing items, while the 28th dimension encodes heel prominence for footwear classes such as boots, sandals, and sneakers.}
	\label{fig:res:fmnist:ours:class-specific}
\end{figure}

To explore class-specific interpretations, we examine the role of latent dimensions 10 and 28. As shown in Figure \ref{fig:res:fmnist:ours:class-specific},the 10th dimension, which is common across upper clothing items, captures garment length. Similarly,the 28th dimension, shared by the "shoes" category—including boots, sandals, and sneakers—encodes the shoe heel. Increasing its value enhances the prominence of the heel . These class-specific interpretations highlight a key advantage of our method, providing insights that disentanglement-based representation learning methods fail to capture. We provide additional experimental results in Appendix~\ref{appendix:fmnist:local}.

\begin{figure}[!htbp]
	\centering
	\includegraphics[width=0.5\textwidth]{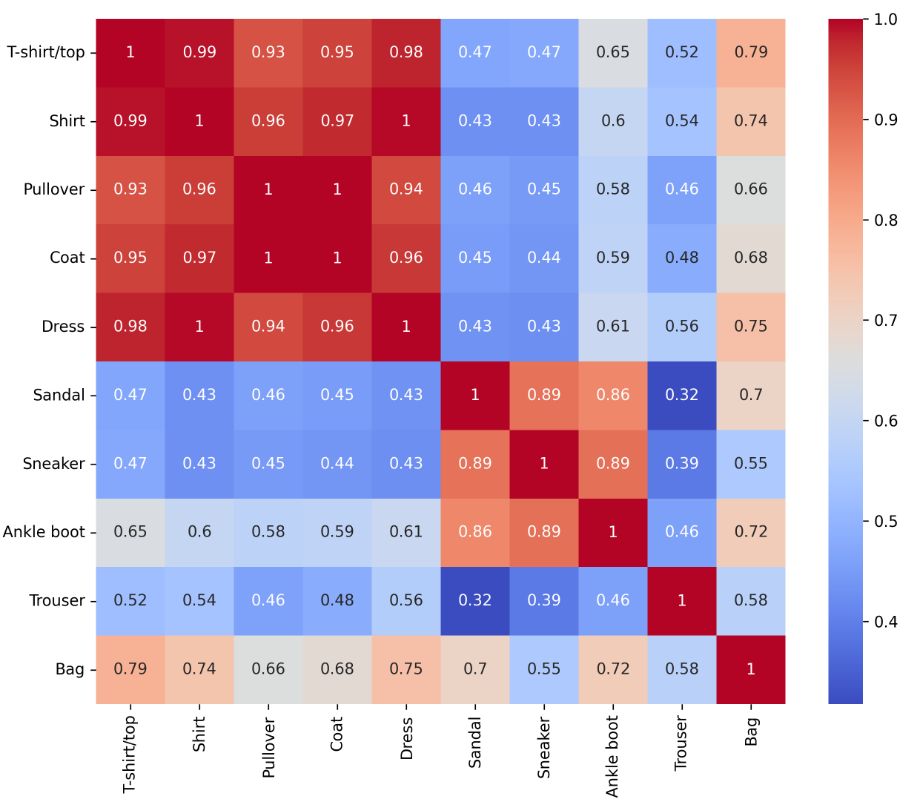}
	\caption{Pearson correlation coefficient between gamma vectors across different Fashion-MNIST classes using our proposed method. Higher correlations occur among classes within the same category.}
	\label{fig:res:fmnist:pearson_coef}
\end{figure}

Beyond individual feature interpretations, it is essential to analyze relationships between sparse vectors. Specifically, we expect that classes within the same category will share more common features and patterns. Figure \ref{fig:res:fmnist:pearson_coef} presents a heatmap of the Pearson correlation coefficient for class pairs. As expected, classes within the same category exhibit high correlations, confirming that our method effectively captures shared features among similar classes. We provide additional experimental results in Appendix~\ref{appendix:fmnist:heatmaps}.

\section{Conclusion}

In this paper, we proposed a novel approach that enhances latent space interpretability by capturing both global and class-specific features. Unlike traditional disentanglement methods, which assume the presence of shared factors across all classes, our method is more adaptable to real-world datasets where common attributes may not exist for all data points. By incorporating class-specific features alongside global ones, we provide a more comprehensive understanding of the latent space. 

Furthermore, our results demonstrate that classes within the same category exhibit stronger shared attributes, reinforcing the structured nature of our approach. This is particularly important in datasets where global attributes are difficult to define uniformly across all classes. In such cases, our method ensures that meaningful latent factors are still captured at the class level, providing interpretable representations even when universal attributes are absent. This dual-level interpretability, both global and class-wise, enhances the practical utility of our method, making it well-suited for applications requiring nuanced insights across and within different classes.

\newpage
\bibliographystyle{unsrtnat}
\bibliography{references}  






\newpage

\appendix
\section*{Appendix}
This appendix provides supplementary results and analysis to support the findings in the paper. It includes extended visualizations, class-specific interpretations, and quantitative evaluations for both MNIST and Fashion-MNIST datasets.

\section{MNIST}
\subsection{MNIST: Global Interpretations}
\label{appendix:mnist:global}
\begin{figure}[H]
	\centering
	\includegraphics[width=0.35\textwidth]{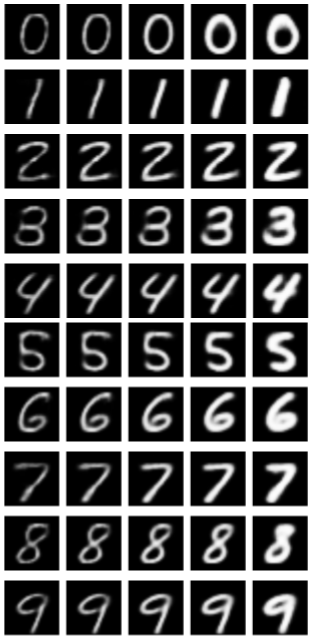}
	\caption{Global feature for the MNIST dataset: The second latent dimension controls digit thickness.}
	\label{fig:res:mnist:ours:globals_appendix_m1}
\end{figure}

\begin{figure}[H]
	\centering
	\includegraphics[width=0.37\textwidth]{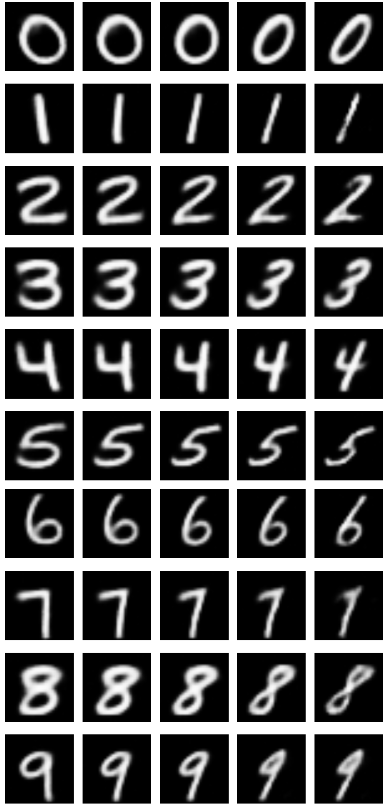}
	\caption{Global feature for the MNIST dataset: The eighteenth latent dimension controls rotation along the $y=x$ axis.}
	\label{fig:res:mnist:ours:globals_appendix_m2}
\end{figure}

\begin{figure}[H]
	\centering
	\includegraphics[width=0.27\textwidth]{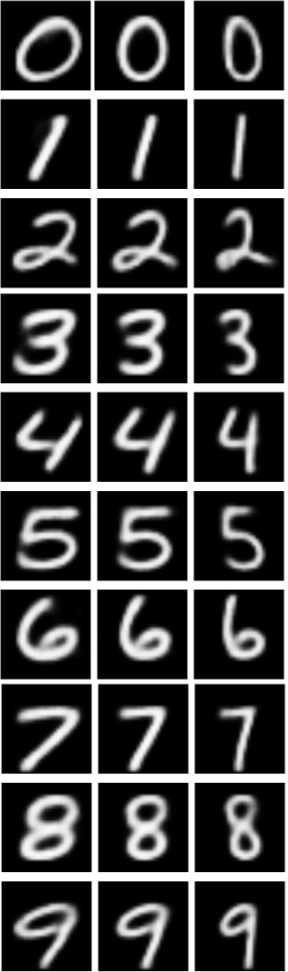}
	\caption{Global feature for the MNIST dataset: The twenty-second latent dimension controls vertical contraction of digits.}
	\label{fig:res:mnist:ours:globals_appendi_m3}
\end{figure}

We analyze latent dimensions that remain active across all digit classes to confirm the presence of global features. These features are consistent with concepts such as stroke thickness, digit slant, and vertical scaling.  
Figure~\ref{fig:res:mnist:ours:globals_appendix_m1} shows how increasing the second latent dimension affects digit thickness, while Figure~\ref{fig:res:mnist:ours:globals_appendix_m2} demonstrates a consistent rotational transformation. Figure~\ref{fig:res:mnist:ours:globals_appendi_m3} illustrates vertical contraction controlled by the twenty-second dimension.

\subsection{MNIST: Class-wise Interpretations}
\label{appendix:mnist:local}
\begin{figure}[H]
	\centering
	\includegraphics[width=0.32\textwidth]{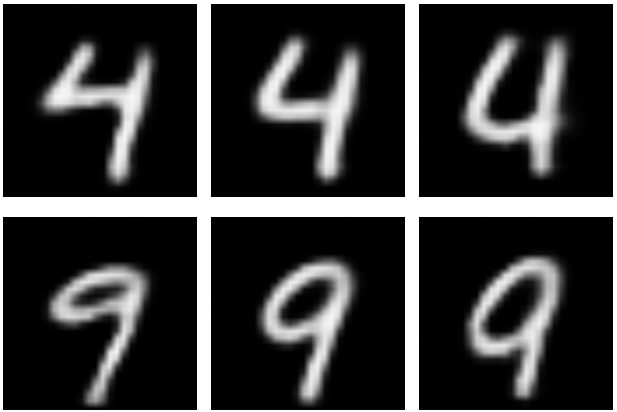}
	\caption{Class-specific feature in the MNIST dataset: The sixth latent dimension controls the intersection position for digits 4 and 9.}
	\label{fig:res:mnist:ours:class-specific_appendix_m1}
\end{figure}

\begin{figure}[H]
	\centering
	\includegraphics[width=0.3\textwidth]{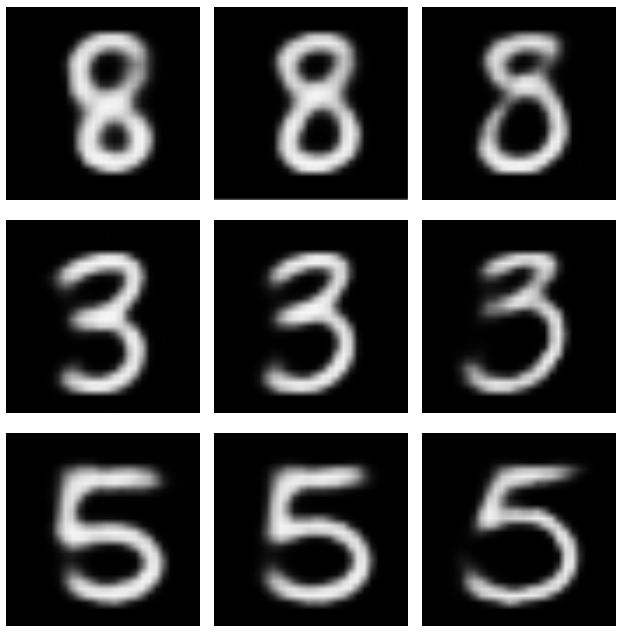}
	\caption{Class-specific feature in the MNIST dataset: The twentieth latent dimension adjusts the lower circle size in digits 3, 5, and 8.}
	\label{fig:res:mnist:ours:class-specific_appendix_m2}
\end{figure}

Certain latent dimensions are active only for specific digits, enabling fine-grained interpretability. For example, Figure~\ref{fig:res:mnist:ours:class-specific_appendix_m1} highlights how the sixth dimension shifts the intersection point in digits 4 and 9. Figure~\ref{fig:res:mnist:ours:class-specific_appendix_m2} shows that the twentieth dimension controls the size of the lower circle in digits 3, 5, and 8. These dimensions do not generalize across all digits, underscoring their class-specific nature.

\begin{figure}[H]
	\centering
	\includegraphics[width=0.3\textwidth]{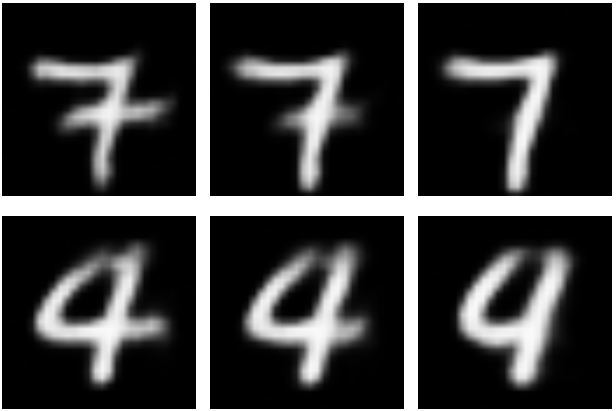}
	\caption{Class-specific feature in the MNIST dataset: The seventh latent dimension affects curvature and shape in specific digit classes.}
	\label{fig:res:mnist:ours:class-specific_appendix_m3}
\end{figure}

\subsection{MNIST: Metrics(Heatmaps)}
\label{appendix:mnist:heatmaps}
\begin{figure}[H]
	\centering
	\includegraphics[width=0.5\textwidth]{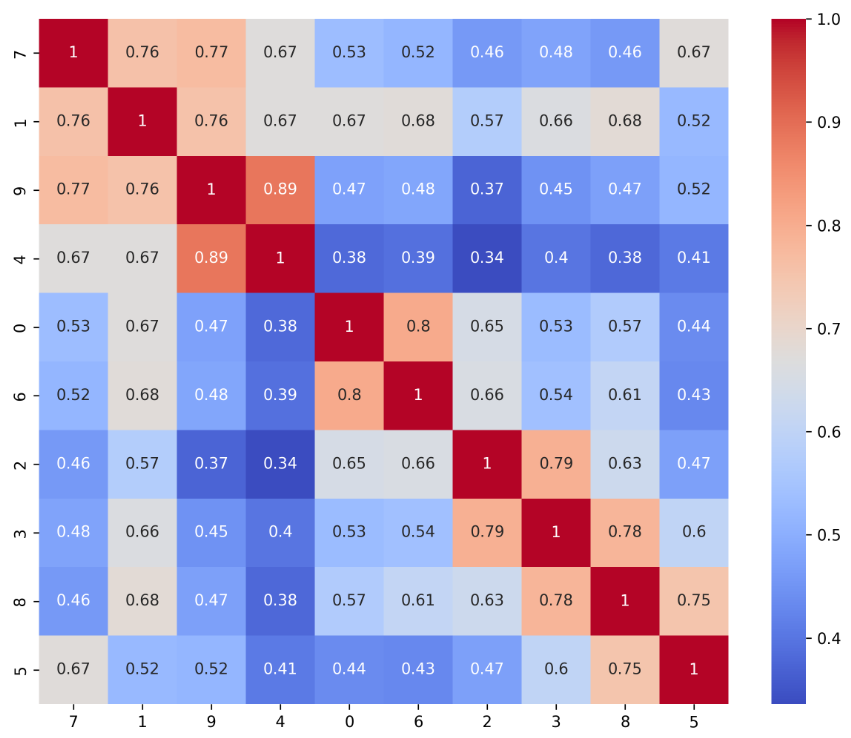}
	\caption{Pearson correlation coefficient between gamma vectors across different MNIST classes using the proposed method.}
	\label{fig:res:mnist:pearson_coef_appendix_m1}
\end{figure}

\begin{figure}[H]
	\centering
	\includegraphics[width=0.5\textwidth]{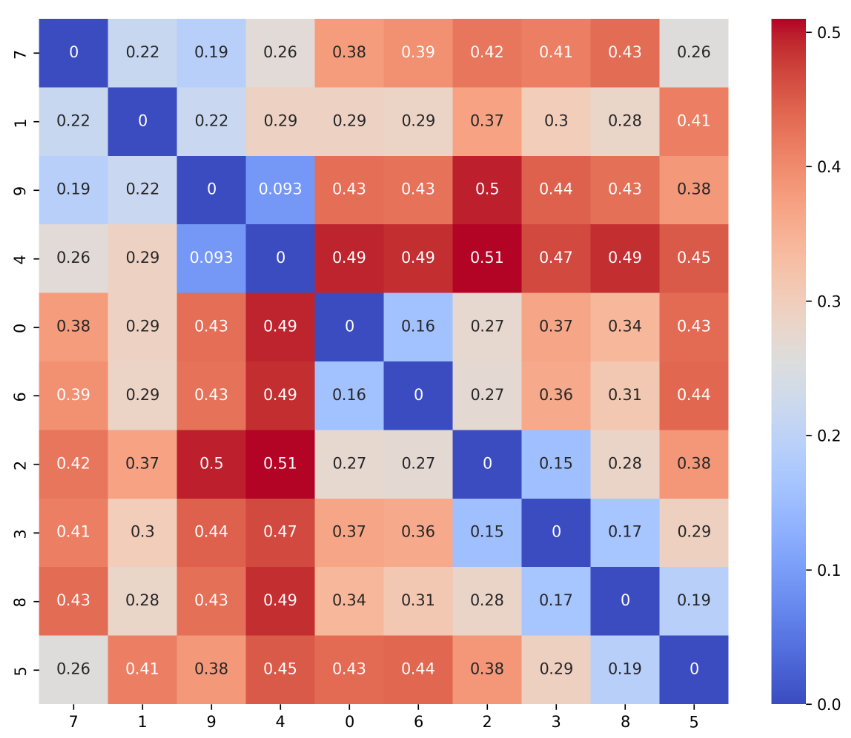}
	\caption{Cosine distance between gamma vectors across different MNIST classes using the proposed method. Lower values indicate stronger similarity.}
	\label{fig:res:mnist:cos_dist_appendix_m2}
\end{figure}

\begin{figure}[H]
	\centering
	\includegraphics[width=0.5\textwidth]{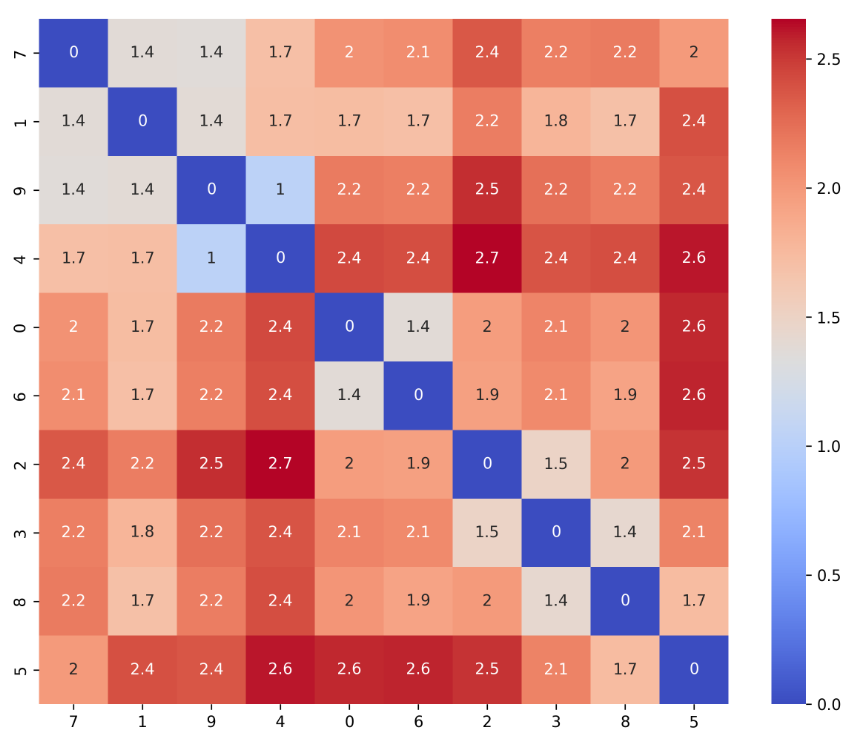}
	\caption{Euclidean distance between gamma vectors across different MNIST classes using the proposed method. Lower values indicate stronger similarity.}
	\label{fig:res:mnist:euclid_dist_appendix_m3}
\end{figure}

To assess latent space consistency, we compare average gamma vectors between classes using different distance metrics. Pearson correlation (Figure~\ref{fig:res:mnist:pearson_coef_appendix_m1}) indicates stronger similarity between structurally similar digits. Cosine (Figure~\ref{fig:res:mnist:cos_dist_appendix_m2}) and Euclidean distances (Figure~\ref{fig:res:mnist:euclid_dist_appendix_m3}) confirm that intra-category distances are lower, validating the model’s structure.

\subsection{MNIST: Metrics(Plots)}

To understand how the model evolves during training, we track two primary metrics: the Jensen-Shannon divergence loss (\(L_{\text{JSD}}\)) and the ELBO.

\begin{figure}[H]
	\centering
	\includegraphics[width=0.8\textwidth]{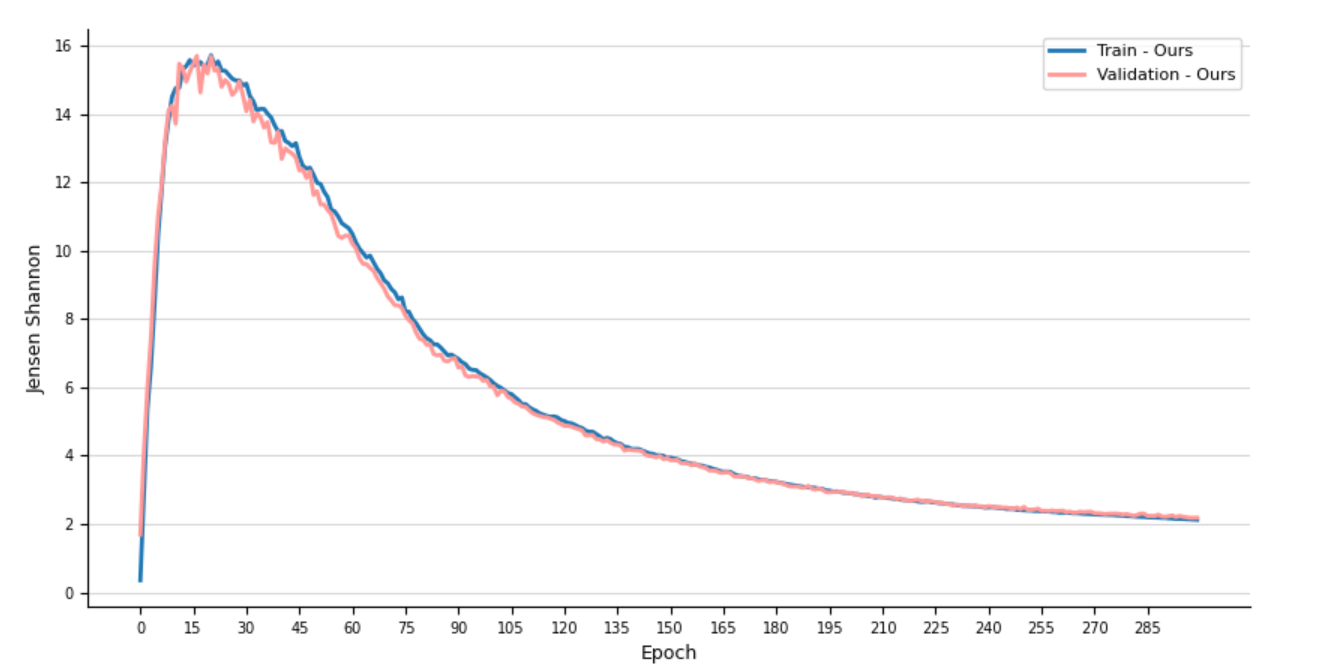}
	\caption{Jensen-Shannon divergence between gamma vectors across training epochs for the MNIST dataset, showing increasing latent alignment.}
	\label{fig:res:mnist:appendix_js}
\end{figure}

Figure~\ref{fig:res:mnist:appendix_js} shows the average Jensen-Shannon divergence between gamma vectors within each class over training epochs. As training progresses, this divergence decreases, indicating that samples from the same class increasingly activate similar latent dimensions. A scheduler is used to gradually increase the weight of this loss term during training.

\begin{figure}[H]
	\centering
	\includegraphics[width=0.8\textwidth]{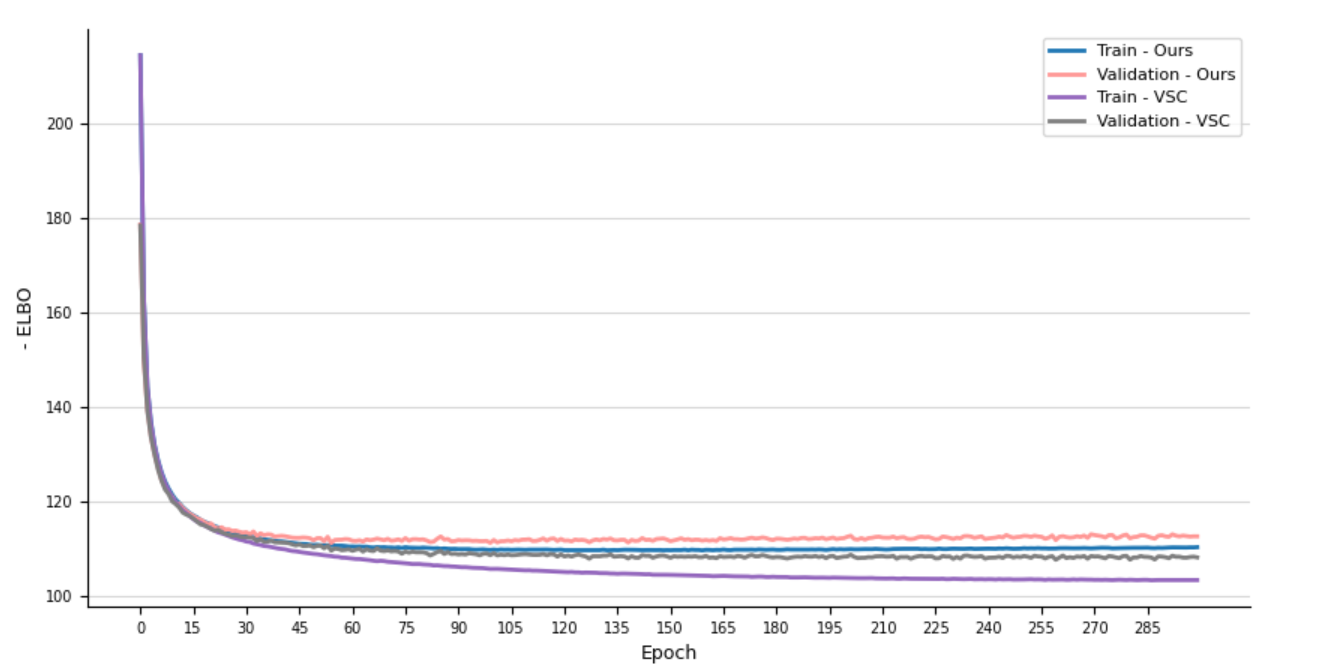}
	\caption{-ELBO values across training epochs for the MNIST dataset.}
	\label{fig:res:mnist:appendix_elbo}
\end{figure}

Figure~\ref{fig:res:mnist:appendix_elbo} plots the negative ELBO, which decreases over time, reflecting improved reconstruction and a better posterior approximation.

\section{Fashion-MNIST}
\subsection{Fashion-MNIST: Class-wise Interpretations}
\label{appendix:fmnist:local}
\begin{figure}[H]
	\centering
	\includegraphics[width=0.4\textwidth]{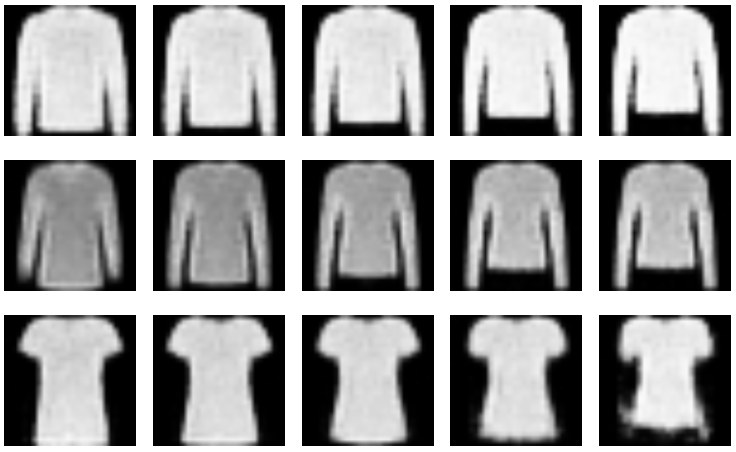}
	\caption{Class-specific feature in the Fashion-MNIST dataset: The tenth latent dimension captures garment length in upper clothing classes.}
	\label{fig:res:fmnist:ours:class-specific_appendix_f1}
\end{figure}

\begin{figure}[H]
	\centering
	\includegraphics[width=0.4\textwidth]{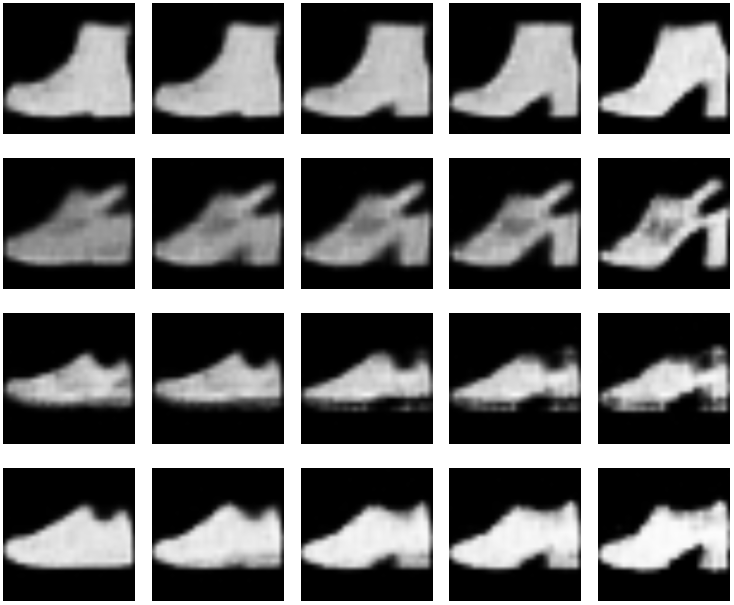}
	\caption{Class-specific feature in the Fashion-MNIST dataset: The twenty-eighth latent dimension controls heel prominence for shoe classes.}
	\label{fig:res:fmnist:ours:class-specific_appendix_f3}
\end{figure}

Latent dimensions in the Fashion-MNIST dataset reveal rich, interpretable patterns. Figure~\ref{fig:res:fmnist:ours:class-specific_appendix_f1} shows that dimension 10 controls clothing length. Figure~\ref{fig:res:fmnist:ours:class-specific_appendix_f3} demonstrates that dimension 28 controls shoe heel prominence. These features vary across categories and enable high-level interpretability of the model’s internal representations.

\begin{figure}[H]
	\centering
	\includegraphics[width=0.4\textwidth]{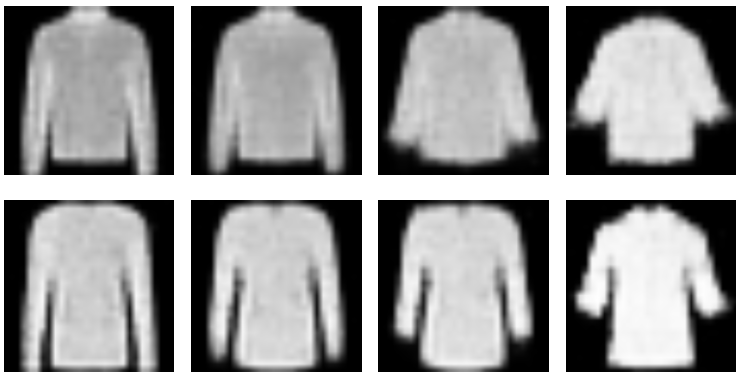}
	\caption{Class-specific feature in the Fashion-MNIST dataset: The nineteenth latent dimension adjusts width or tightness for some clothing classes.}
	\label{fig:res:fmnist:ours:class-specific_appendix_f2}
\end{figure}

\begin{figure}[H]
	\centering
	\includegraphics[width=0.5\textwidth]{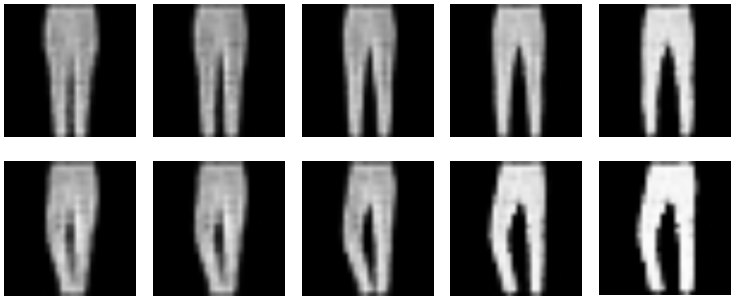}
	\caption{Class-specific feature in the Fashion-MNIST dataset: The thirtieth latent dimension affects the sole thickness or elevation in footwear.}
	\label{fig:res:fmnist:ours:class-specific_appendix_f4}
\end{figure}

\begin{figure}[H]
	\centering
	\includegraphics[width=0.4\textwidth]{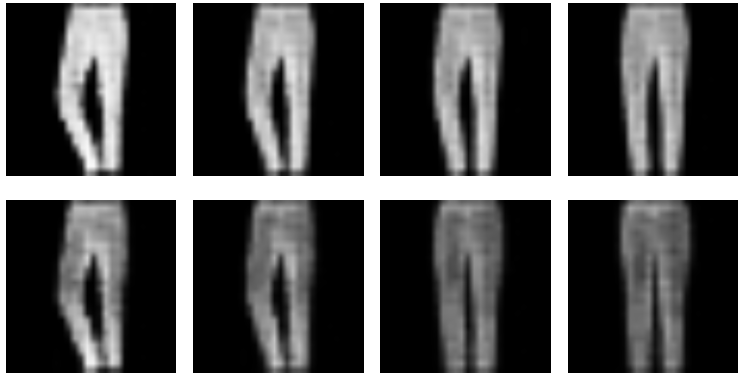}
	\caption{Class-specific feature in the Fashion-MNIST dataset: The eighth latent dimension adjusts shape details for upper garments.}
	\label{fig:res:fmnist:ours:class-specific_appendix_f5}
\end{figure}

\subsection{Fahion-MNIST: Metrics(Heatmaps)}
\label{appendix:fmnist:heatmaps}
\begin{figure}[H]
	\centering
	\includegraphics[width=0.55\textwidth]{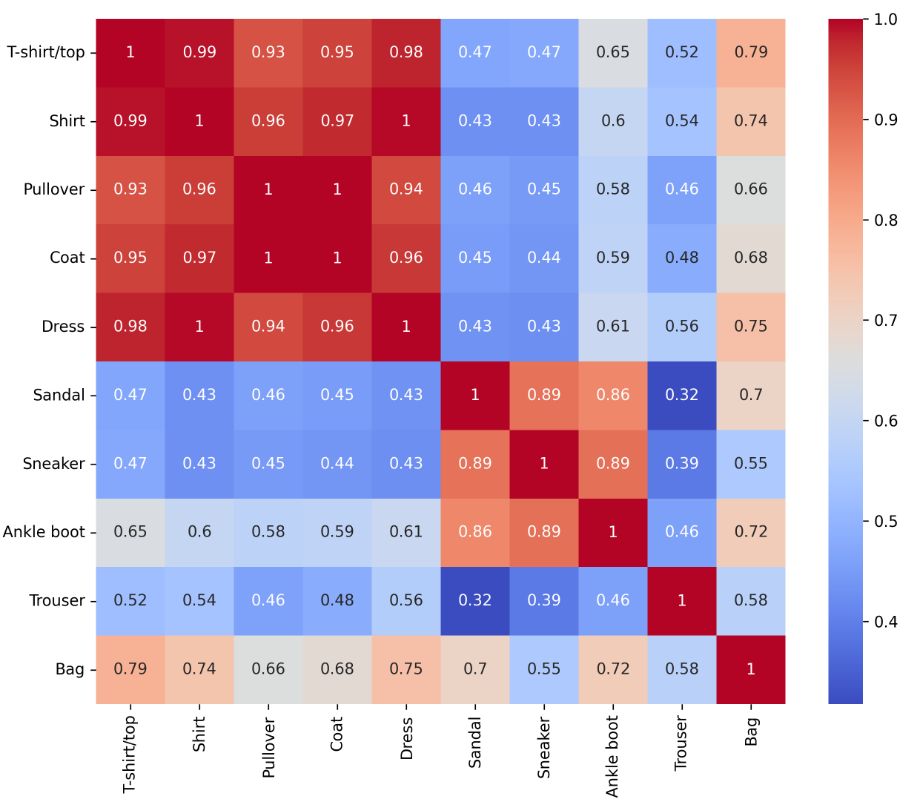}
	\caption{Pearson correlation coefficient between gamma vectors across different Fashion-MNIST classes using the proposed method.}
	\label{fig:res:mnist:pearson_coef_appendix_f1}
\end{figure}

\begin{figure}[H]
	\centering
	\includegraphics[width=0.55\textwidth]{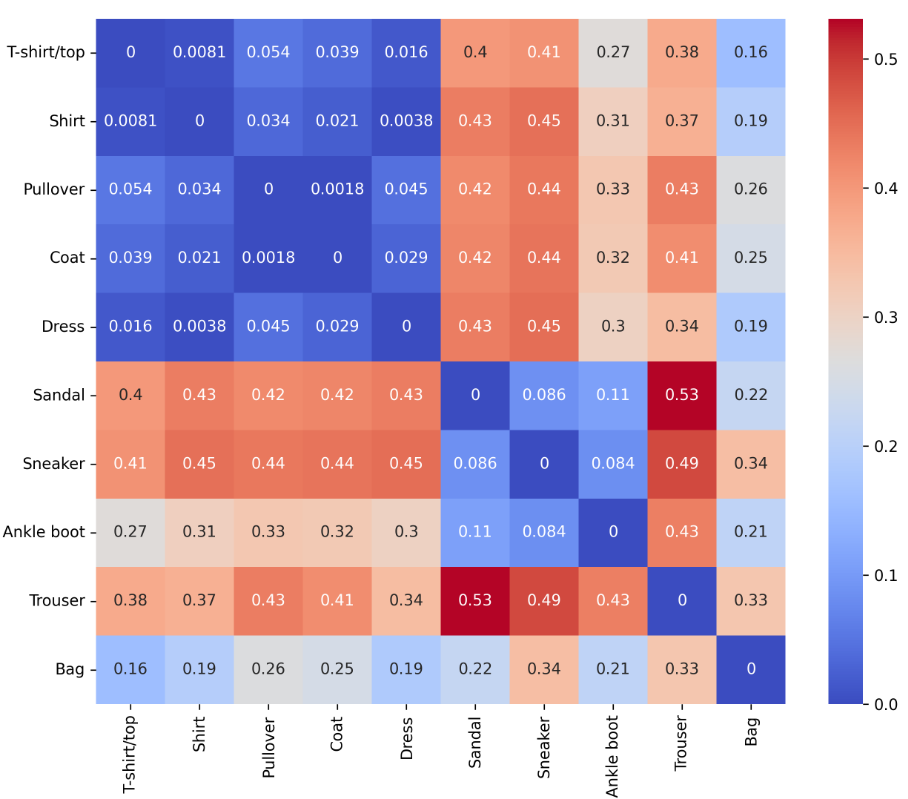}
	\caption{Cosine distance between gamma vectors across different Fashion-MNIST classes using the proposed method.}
	\label{fig:res:fmnist:cos_dist_appendix_f2}
\end{figure}

\begin{figure}[H]
	\centering
	\includegraphics[width=0.55\textwidth]{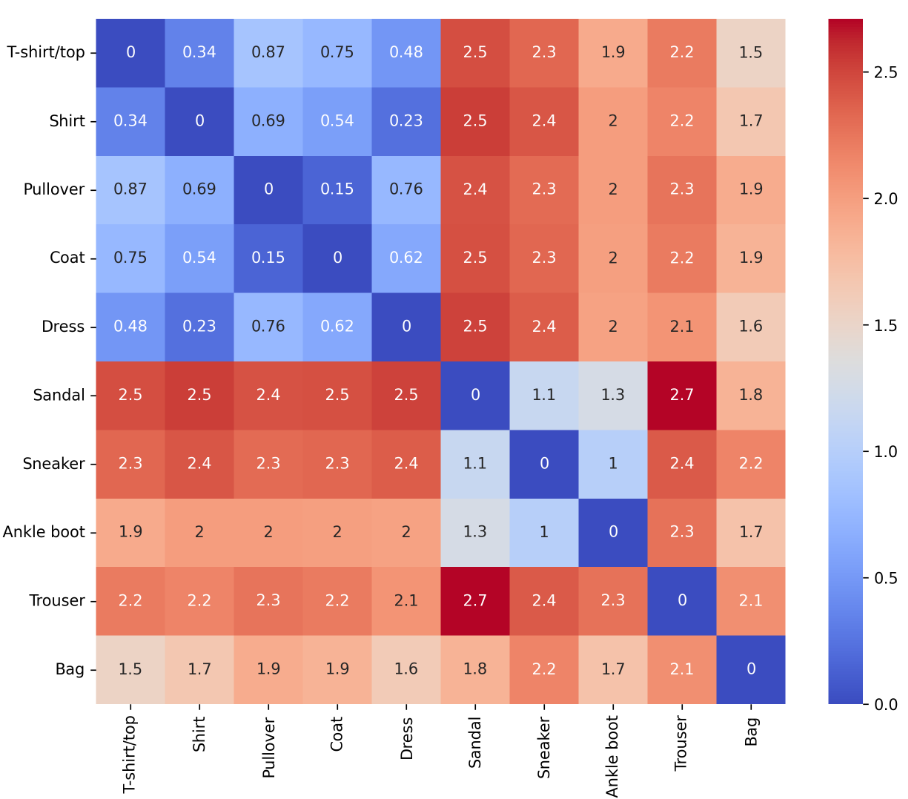}
	\caption{Euclidean distance between gamma vectors across different Fashion-MNIST classes using the proposed method.}
	\label{fig:res:fmnist:euclid_dist_appendix_f3}
\end{figure}

We present gamma vector similarities across Fashion-MNIST classes. As shown in Figure~\ref{fig:res:mnist:pearson_coef_appendix_f1}, classes such as boots, sandals, and sneakers exhibit high Pearson correlation, indicating shared latent structure. Similar trends are confirmed using cosine and Euclidean distances (Figures~\ref{fig:res:fmnist:cos_dist_appendix_f2},~\ref{fig:res:fmnist:euclid_dist_appendix_f3}).

\subsection{Fashion-MNIST: Metrics(Plots)}
\begin{figure}[H]
	\centering
	\includegraphics[width=0.8\textwidth]{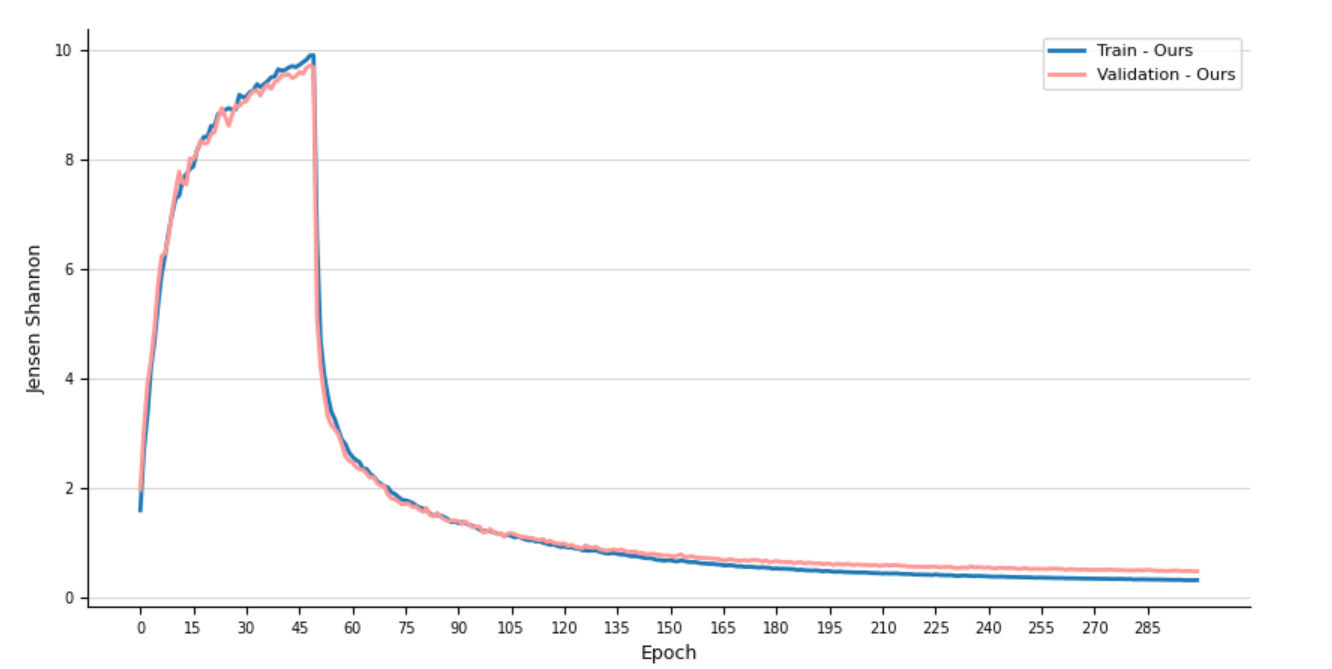}
	\caption{Jensen-Shannon divergence during training for the Fashion-MNIST dataset. Lower values indicate increased alignment of active latent dimensions.}
	\label{fig:res:fmnist:appendix_js}
\end{figure}

\begin{figure}[H]
	\centering
	\includegraphics[width=0.8\textwidth]{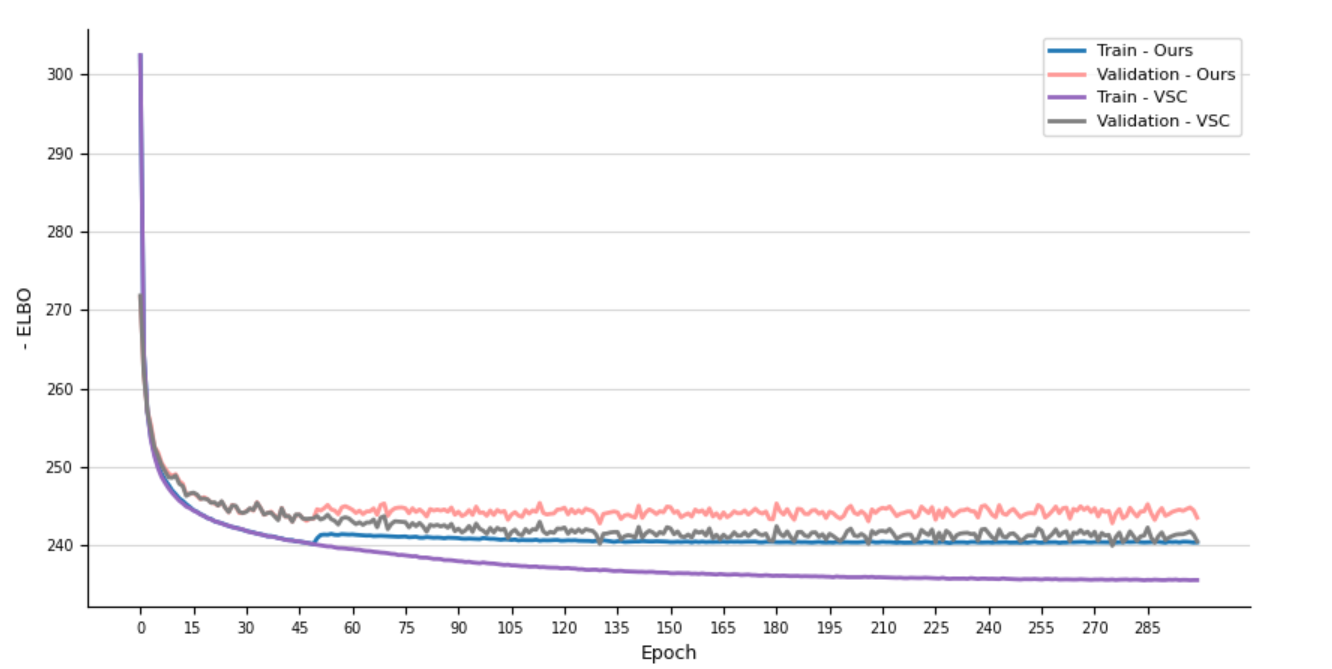}
	\caption{Negative ELBO values during training for the Fashion-MNIST dataset. Lower negative ELBO values correspond to better model performance.}
	\label{fig:res:fmnist:appendix_elbo}
\end{figure}

Figure~\ref{fig:res:fmnist:appendix_js} shows a steady decline in divergence, beginning after epoch 45 when the loss term is applied, indicating improved alignment of active latent dimensions within each class. Figure~\ref{fig:res:fmnist:appendix_elbo} shows a decreasing negative ELBO, reflecting improved reconstruction and posterior approximation.

\end{document}